\documentclass{article}

\PassOptionsToPackage{numbers,compress,sort}{natbib}

\usepackage[final]{main}

\usepackage[utf8]{inputenc} 
\usepackage[T1]{fontenc}    
\usepackage{hyperref}       
\usepackage{url}            
\usepackage{booktabs}       
\usepackage{amsfonts}       
\usepackage{nicefrac}       
\usepackage{microtype}      
\usepackage{xcolor}         
\usepackage{caption}

\usepackage{appendix}
\usepackage{fontawesome5}
\usepackage{mathtools}
\usepackage{amsfonts,amsmath,amsthm}
\usepackage[capitalise,noabbrev]{cleveref}
\usepackage{tabularx}
\AtBeginEnvironment{tcolorbox}{\small}

\usepackage{makecell}
\usepackage{enumitem}
\usepackage{xcolor}
\usepackage{multirow}
\usepackage{tcolorbox}
\usepackage{algorithm}
\usepackage{wrapfig}
\usepackage{algpseudocode}
\usepackage{newunicodechar}
\newunicodechar{χ}{$\chi$}

\usepackage{graphicx}

\newcommand{\GemPro}{\text{Gemini-2.5-Pro}}
\newcommand{\GemFlash}{\text{Gemini-2.0-Flash}}
\newcommand{\Llama}{\text{Llama-3.3-70B}}
\newcommand{\QwQ}{\text{Qwen-QwQ-32B}}
\newcommand{\Qwen}{\text{Qwen3-32B}}
\newcommand{\othree}{\text{o3-mini}}
\newcommand{\DSD}{\text{DeepSeek-70B}}
\newcommand{\Deepseek}{\text{DeepSeek-R1}}
\newcommand{\DSLlamaSmall}{\text{DeepSeek-8B}}
\newcommand{\DSQwenMed}{\text{DeepSeek-14B}}

\renewcommand{\>}{\succ}

\newtheorem{example}{Example}

\newtheorem{result}{Finding}

\title{Matching Markets Meet LLMs: \\Algorithmic Reasoning with Ranked Preferences}

\author{
\textbf{Hadi Hosseini} \\ 
Penn State University, USA\\ 
\texttt{hadi@psu.edu}
\and 
\textbf{Samarth Khanna} \\
Penn State University, USA\\ 
\texttt{samarth.khanna@psu.edu}
\and
\textbf{Ronak Singh} \\ 
Penn State University, USA\\ 
\texttt{ronak.singh@psu.edu}
}

\begin{document}

\maketitle

\begin{abstract}

The rise of Large Language Models (LLMs) has driven progress in reasoning tasks, from program synthesis to scientific hypothesis generation, yet their ability to handle ranked preferences and structured algorithms in combinatorial domains remains underexplored. 
We study matching markets, a core framework behind applications like resource allocation and ride-sharing, which require reconciling individual ranked preferences to ensure stable outcomes. 
We evaluate seven state‐of‐the‐art models on a hierarchy of preference‐based reasoning tasks---ranging from stable‐matching generation to instability detection, instability resolution, and fine-grained preference queries---to systematically expose their logical and algorithmic limitations in handling ranked inputs.
Surprisingly, even top-performing models with advanced reasoning struggle to resolve instability in large markets, often failing to identify blocking pairs or execute algorithms iteratively.  
We further show that \textit{parameter-efficient fine-tuning} (LoRA) significantly improves performance in small markets, but fails to bring about a similar improvement in large instances, suggesting the need for more sophisticated strategies to improve LLMs' reasoning with larger-context inputs. 
\end{abstract}
\begin{center}
    \faGithub\ \href{https://github.com/SamarthKhanna/LLM_Matching_Markets}{\texttt{Data and Code: github.com/SamarthKhanna/LLM\_Matching\_Markets}}
\end{center}

\section{Introduction}

The emergence of Large Language Models (LLMs) has positioned them as integral components in a wide range of reasoning-intensive tasks such as program synthesis, logical inference, mathematical problem solving, and scientific hypothesis generation, highlighting the importance of structured problem-solving capabilities.
Despite their recent success in symbolic and logical reasoning, their capacity to reason over ranked preferences and to execute structured algorithms within combinatorial domains remains largely unexplored.
Preference reasoning constitutes a foundational component in numerous domains, including economic contexts---e.g., auctions, voting systems, and market design---and in the architecture of pre-trained generative models using Reinforcement Learning from Human Feedback (RLHF) to capture and internalize human value judgments.
These methods often have to execute algorithms on a large number of preference lists (either pairwise, partial, or complete rankings) to aggregate the rankings through constitutional AI \citep{bai2022constitutional} or social choice theory \citep{Conitzer+24}.

Despite substantial progress, reasoning over preferences remains a non-trivial endeavor: ensuring transitivity \citep{xu2025investigating,song2025benchmarking}, accurately augmenting ordinal rankings \citep{fish2023generative}, and achieving coherent value alignment pose significant challenges. 
Without robust mechanisms for preference elicitation and the capacity to execute the requisite combinatorial procedures, even state-of-the-art LLMs may produce outputs that diverge from true human preferences \citep{hosseini2025distributive} or fail to satisfy desirable properties \citep{Fish2025EconEvals}.

We consider matching markets, a domain that constitutes a fundamental class of problems underlying diverse applications—from healthcare resource allocation to ride-sharing platforms and recommender systems—and demand accurate comprehension of individual preferences and reconciliation of conflicting choices to guarantee system-wide stability.
Matching markets are an ideal testbed for studying reasoning in AI models for two key reasons:
First, they provide a structured platform for evaluating reasoning over ranked preferences and algorithmic thinking. They provide a framework that is axiomatically rich yet computationally tractable where solution quality (e.g., stability and efficiency) can be rigorously evaluated.
Second, LLMs are increasingly utilized as black-box systems in a variety of economic, social, or medical settings to inform automated screening in recruitment pipelines \citep{gan2024application},
investigating market behavior \citep{jia2024experimental}, market clearing in ride-hailing platforms \citep{junque2025simulating}, and in general simulating economic interactions \citep{horton2023large}. This makes it imperative to evaluate their ability in terms of computing ``desirable'' solutions from stakeholders' preferences. Additionally, their potential to serve as \textit{interfaces} between stakeholders and established black-box systems (e.g., the National Residents Matching Program \citep{NRMP2024}) requires benchmarking their ability to reason about provided solutions and address aspects that users might find undesirable.

\subsection{Our Results}\label{sec:results}

\begin{figure}
    \centering
    \includegraphics[width=0.95\linewidth]{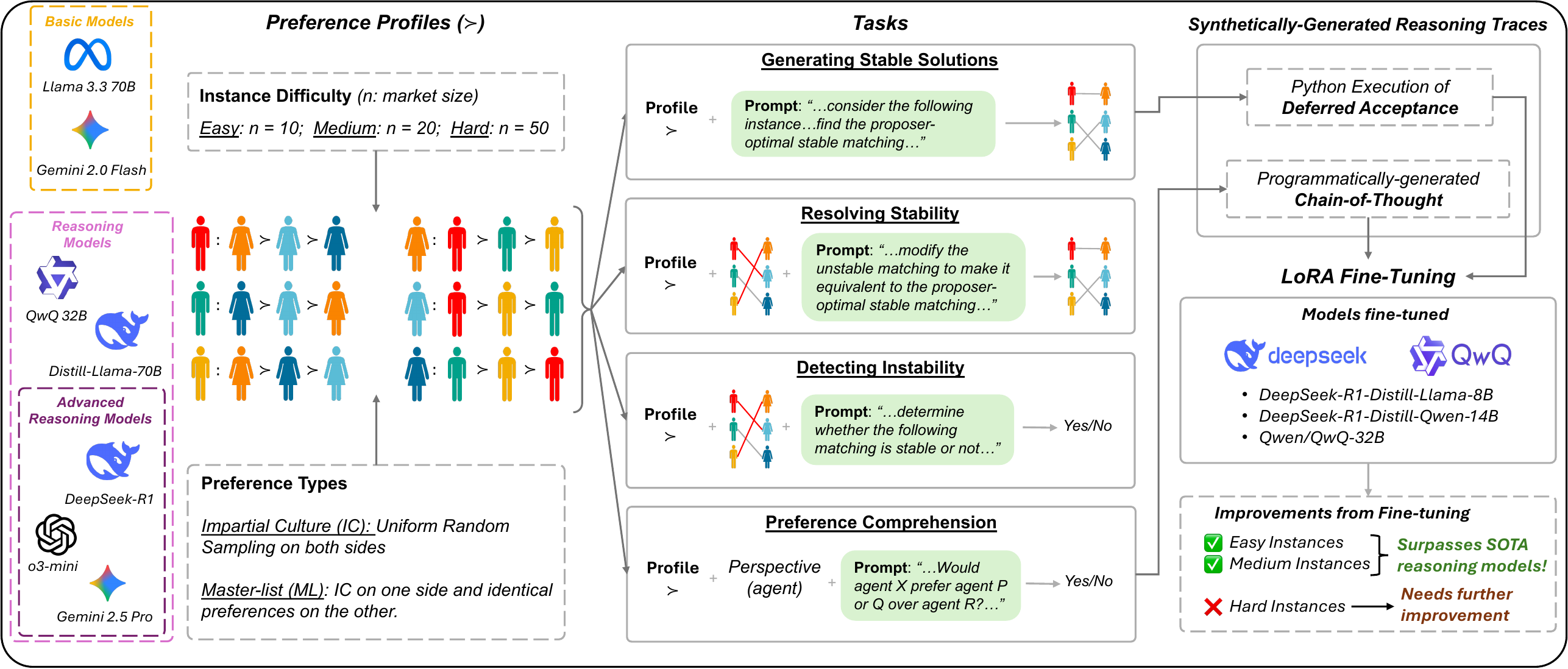}
    \caption{The framework for reasoning with ranked preferences through matching markets.}
    \label{fig:teaser}
\end{figure}

We focus on four preference-based tasks:
(i) \textit{generating} stable solutions, requiring LLMs to produce stable matchings directly from ranked inputs,
(ii) \textit{instability resolution}, demanding preference reasoning to transform unstable matchings to stable ones,
(iii) \textit{instability detection}, in which models detect blocking pairs within a proposed solution, and 
(iv) \textit{preference reasoning}, assessing nuanced query answering over ranked lists.
We evaluate seven large language models with varying reasoning capabilities, including basic models, those with some limited reasoning, and advanced reasoning models.
Our methodology and results are summarized in \Cref{fig:teaser}.

\textbf{Benchmark.}
We introduce a benchmark with instances and questions aimed at evaluating the above tasks involving reasoning over ranked preferences. 
These tasks are categorized into three levels of difficulty—Easy, Medium, and Hard—based on problem size. Each task utilizes ranked preferences sampled from two statistical distribution models: Impartial Culture (IC) and Master-list (ML).

\textbf{Generating Stable Solutions.} Although models with advanced-reasoning capabilities generally outperform other LLMs on Easy and Medium instances, \textit{all} models struggle to generate stable solutions on Hard instances---indicating that the combinatorial reasoning capability of LLMs does not necessarily extend to larger-context inputs. Interestingly, the fraction of invalid and failed solutions is significantly lower for models with higher reasoning abilities, indicating their understanding of constraints, despite their inability to perform precise and step-by-step reasoning with preferences.

\textbf{Instability Detection and Resolution.} We find that LLMs frequently make mistakes in determining whether solutions are stable, with hallucinations about blocking pairs being the most common among basic models. Additionally, LLMs' ability to \textit{correct} unstable solutions is (at best) as good as their ability to generate them from scratch, in some cases making the provided incorrect solutions worse.

\textbf{Preference Reasoning.}
We consider tasks based on three levels of inference over ranked preferences.
Large language models with advanced reasoning capabilities generally demonstrate a strong comprehension of preferences across levels of inference. 
However, even small errors compound in tasks requiring multi-step sequences of reasoning (e.g., generating stable solutions or resolving instability), or in other words, small errors multiply!

\textbf{Supervised Parameter-Efficient Fine-Tuning.} We demonstrate that fine-tuning an open-source reasoning model using synthetically generated reasoning traces substantially improves performance, significantly outperforming advanced-reasoning models on Easy and Medium instances. However, we find that this approach does not address the challenges LLMs face with large inputs (Hard).

\subsection{Related Work}

\textbf{Reasoning Capabilities of LLMs.} 
Mathematical problem solving has been a key area of focus in evaluating the reasoning ability of LLMs, through a variety of benchmarks such as  \citep{HendrycksMATH2021,HendrycksMMLU2021,CobbeGSM8K2021,Mirzadeh2024GSMSymbolic}. LLMs have also demonstrated remarkable capabilities on coding benchmarks such as SWE-Bench \citep{Jiminez2024SWEBench} and CodeForces. As SOTA benchmark scores improve, recent work studies whether these improvements reflect genuine logical reasoning through benchmarks assessing logical consistency \citep{Liu2024Logic} and rule understanding/execution/planning \cite{Gui2024Rule}. Furthermore, the recent rise of LLM agents has increased interest in benchmarking LLMs' causal reasoning \citep{Haoang2024Causal} and strategic planning abilities \citep{Duan2024GTBench,Tang2025DSGBench,Jia2024Strategic}. Additionally, the emergence of reasoning models has led to benchmarks evaluating these models' improved reasoning and planning abilities \citep{ChenReasoningEra2025,LiSystem22025}.

\textbf{Enhancing Reasoning Capabilities of LLMs.} Specialized prompting strategies like Chain-of-Thought (CoT) \citep{wei2022chain}, Tree-of-Thought (ToT) \citep{Yao2023ToT}, and Graph-of-Thought (GoT) \citep{Besta2024GoT} have performance abilities on a variety of reasoning benchmarks. Fine-tuning has also been demonstrated to improve CoT in model outputs \citep{Zhang2024Chain}, as well as economic rationality \citep{chen2023emergence} and abstract reasoning \citep{Globersons2024Learning}. Additionally, instruction-tuning has been shown to enhance reasoning in several works \citep{Luo2024Improve,Vaillancourt_2024,Cai2024System2,Li2024MuggleMath}. More advanced techniques build upon CoT \citep{Tian2024Improvement,Zhou2024Discover}, or utilize multi-agent architectures that leverage cooperative LLMs \citep{Zhou2024Reso, Trencsenyi2025Reasoning}. More recently, reinforcement-learning (GRPO) has been used to improve model reasoning, the most popular example being the Deepseek-R1 reasoning model \cite{DSR1}.

\textbf{LLMs in Social and Economic Decision Making.} While still an emerging area of research, multiple works have focused on the collective decision-making capabilities of LLMs. One particular area of interest is the use of LLMs in preference elicitation \citep{Soumalis2025Elicitation, Huang2025Accelerated}. \citet{Fish2025EconEvals} benchmark the ability of models to learn and strategize in unknown economic environments using deliberate exploration. Another notable avenue of work is the study of how well LLMs can represent humans in collective decision-making, an understudied component of LLM alignment \citep{Yang2024Voting,hosseini2025distributive}.

\section{Methodology}

\subsection{Problem Formulation} \label{sec:problem}
A \textit{two-sided matching market} consists of two disjoint sets of agents (e.g., riders and drivers, freelancers and job requesters, and content creators and ads) denoted by $M$ and $W$, where $|M| = |W| = n$.
The preference list of an agent $i$, denoted by $\>_i$, is a ranked order list over the agents on the other side. A \textit{preference profile}, $\>$, denotes the collection of preferences of all agents. 
We write $w_1 \succ_m w_2$ and $m_1 \succ_w m_2$ to denote that $m$ \textit{prefers} $w_1$ to $w_2$ and $w$ \textit{prefers} $m_1$ to $m_2$ respectively.
In this paper, we primarily consider the standard model, which assumes a complete and strict preference list (no ties) and aims at finding a \textit{one-to-one} matching between the agents in two sets.\footnote{This is the standard model considered by the seminal works of \citet{gale1962college} and \citet{knuth1997stable}.}

\textbf{Matching and Stability.}
A \emph{matching} is a function $\mu: M \cup W \rightarrow M \cup W$ such that $\mu(m) \in W$ for all $m \in M$, $\mu(w) \in M$ for all $w \in W$, and $\mu(m) = w$ if and only if $\mu(w) = m$. 
Given a matching $\mu$, a \emph{blocking pair} with respect to the preference profile $\>$ is a pair $(m, w)$ who prefer each other over their assigned partners in $\mu$, i.e., $w \>_m \mu(m)$ and $m \>_w \mu(w)$. A matching is said to be \emph{stable} if it does not have any blocking pairs.
Given an instance of the problem, the set of all possible stable solutions forms a distributive lattice and can be \textit{exponential} in size \citep{knuth1997stable}.

In their seminal work, \citet{gale1962college} proposed an iterative procedure---the \textit{deferred acceptance} algorithm (DA)---that always guarantees to find a stable solution. 
It proceeds by a series of proposals and rejections. In the initial \emph{proposal} phase, each of the unmatched agents on one side (aka \textit{proposers}) proposes to their favorite agent from the other side (aka \textit{receivers}) according to their preference list. In the subsequent \emph{rejection phase}, each agent on the receiving side tentatively accepts its preferred proposal, rejecting the others. The algorithm terminates when no further proposals can be made.
The details of this
algorithm can be found in \Cref{app:algorithms}.
The underlined solution in \Cref{ex:example1} is simultaneously \emph{optimal} for the proposing side and \textit{pessimal} for the receiving side \citep{mcvitie1971stable}. We refer to the former as the \textbf{Optimal} matching and the latter as the \textbf{Pessimal} matching.

\begin{example} [An instance with multiple stable solutions.]\label{ex:example1}
A preference profile for a sample instance of size $n=4$; underlined agents indicate the Optimal matching, the Pessimal matching is indicated with a $^*$, and the $\dagger$ indicates a stable matching that is different from the first two.
\begin{table}[h]
\footnotesize
    \centering
    \begin{tabularx}{0.55\linewidth}{XXXXXXXXXXXXXXX}
            ${\boldsymbol{m_1}}\colon$ & $\underline{w_4}$ & $w_3$ & $w_1^{*,\dagger}$ & $w_2$ && $\boldsymbol{w_1}\colon$ & $m_2$ & ${m_1^{*,\dagger}}$ & $\underline{m_3}$ & $m_4$ \\
            ${\boldsymbol{m_2}}\colon$ & $\underline{w_3}$ & ${w_4^{\dagger}}$ & $w_2^*$ & $w_1$ && ${\boldsymbol{w_2}}\colon$ & $m_2^*$ & $m_3$ & ${\underline{m_4^{\dagger}}}$ & $m_1$ \\
            ${\boldsymbol{m_3}}\colon$ & $\underline{w_1}$ & $w_3^{\dagger}$ & $w_2$ & $w_4^*$ && $\boldsymbol{w_3}\colon$ & $m_4^*$ & ${m_3^{\dagger}}$ & $m_1$ & $\underline{m_2}$ \\
            $\boldsymbol{m_4}\colon$ & $w_1$ & ${\underline{w_2^{\dagger}}}$ & $w_3^*$ & $w_4$ && $\boldsymbol{w_4}\colon$ & $m_4$ & $m_3^*$ & ${m_2^{\dagger}}$ & $\underline{m_1}$ 
        \end{tabularx}
\end{table}
\end{example}

\subsection{Dataset, Models, and Setup} \label{sec:profiles}

\textbf{Preference Instances.} 
We synthetically sample a set of $300$ preference profiles, partitioned into three sets of $100$ instances for each \textit{difficulty level}, namely \textbf{Easy} ($n=10$ agents on each side of the market), \textbf{Medium} ($n=20$), and \textbf{Hard} ($n=50$). 
The preference profiles are sampled from two types of distributions \textbf{Impartial Culture} (IC) and \textbf{Master-list} (ML), each constituting $50$ questions at each difficulty level. 
An impartial culture (IC) is a well-studied probabilistic model for generating preference profiles in which every agent's strict preference ranking is drawn independently and uniformly at random \citep{black1958theory,eugeciouglu2013impartial}. It has been extensively studied in the context of economics, matching, and voting theory \citep{brilliantova2022fair,boehmer2024guide, TsetlinIC, Caragiannis2024IC}. 
A profile with a master‐list (ML) is a highly structured preference profile in which all agents on one side of the market share exactly the same strict ranking 
over the agents on the other side. They represent the \textit{homogeneity} in settings ranging from the labor market to organ allocation in healthcare \citep{irving2008stable, Bredereck2021MasterList, NaoyukiMLMatroid} 
While an arbitrary instance generated by IC may admit \textit{exponentially} many stable solutions \citep{knuth1997stable}, with a Master-list, only a single \textit{unique} stable solution exists, indicating a difficulty level proportional to the size of the space of stable outcomes. In \Cref{app:algorithms}, we discuss a simpler version of the DA algorithm for computing stable solutions with ML instances.

\textbf{Matching Dataset.} 
We curate a dataset comprising 2850 questions derived from the instances described above.
These questions cover four \textit{task categories}, each applied to the same pool of profiles to ensure consistency: (i) \textbf{generating stable solutions}, given a preference profile ($300$ questions); (ii) \textbf{instability resolution}, given a profile and an unstable matching; (iii) \textbf{instability detection}, given a profile and a solution ($1050$ questions); and (iv) \textbf{preference reasoning}, given a single preference list or a profile ($900$ questions).

\textbf{Models.} We select a representative suite of both open-source and closed-source models for evaluation. Since our benchmark is based on a reasoning task, we categorize models by their reasoning ability. We evaluate two \textbf{basic} models (those not specifically trained for reasoning), namely \Llama{} \citep{Dubey2023Llama} and \GemFlash{} \citep{sundar_pichai_2024}, and five \textbf{reasoning} models, namely \QwQ{} \citep{qwen_paper}, \DSD{} (Llama-distilled) \citep{DSR1}, OpenAI \othree{} \citep{openai_o3mini_2025}, \Deepseek{} \citep{DSR1}, and \GemPro{} \citep{deepmind_2025}. Among reasoning models, we classify the last three as \textbf{advanced reasoning} models, based on their SOTA performance on reasoning benchmarks \citep{LiSystem22025}. 

\textbf{Prompting.} The prompt for each task consists of the preference profile for a given instance, followed by task-specific instructions (e.g., computing the ``proposer-optimal'' matching, or resolving a given unstable matching). While we adopt the \textit{stable-marriage} setting, considered by \citet{gale1962college}, where \textit{men} propose to \textit{women}, we show (in \Cref{app:CoT}) that the results do not change if a different setting---where \textit{workers} are assigned \textit{tasks}---is used.
To scale up the verification of solution correctness, we instruct LLMs to adhere to a predefined answer format. Additionally, we allow LLMs two \textit{re-tries} to correct solutions that are either invalid, partial, or do not adhere to the specified format. See \Cref{app:inference} for details about the inference setup, and \Cref{app:prompts} for sample prompts.

\subsection{Evaluation Criteria}\label{sec:metrics}
We consider several metrics for evaluating the quality of returned responses depending on the task. 
To account for cases in which LLM outputs violate task requirements, we categorize responses into the following types:
A solution is \textbf{invalid} if \textit{some} agent from one side is matched to more than one agent from the other side. 
It is \textbf{partial} if it is not invalid, but some agents remain unmatched. 
A matching is \textbf{stable} if it is a \textit{perfect} one-to-one matching that admits no blocking pair. Otherwise, it is \textbf{unstable} if it matches all the agents but admits a blocking pair. 
The following metrics apply primarily to valid responses. Informally, these metrics measure the distance from a reference stable outcome.

\textbf{Instability Rate (IR)}: The instability rate measures the proportion of agents involved in blocking pairs, and thus the degree to which a matching violates the stability criterion. 
Given a complete matching, instability rate measures the percentage of unstable agents, i.e., those involved in at least one blocking pair. Formally, 
     $ \text{IR}(\mu, \>) = \tfrac{|\{i\in M\cup W\ \text{s.t.}\ j\ \>_{i} \mu(i)\ \wedge\ i\ \>_{j} \mu(j)\ \text{for some } j\in M\cup W\}|}{2n}.
    $
 
    \textbf{Optimality/Pessimality Rate}: 
    This rate assesses the overlap between the model's matching and a reference stable matching, thereby capturing how closely the model's output mirrors the stepwise proposals and acceptances of a canonical algorithm.
    Formally, given two perfect matchings, $\mu$ and $\mu'$, in a one‐to‐one market where each matching is viewed as a set of unordered pairs between agents, the Jaccard similarity is defined as
    $
    \text{JS}(\mu,\mu') = \frac{|\mu \cap \mu'|}{|\mu \cup \mu'|}
    $. 
    Then, we define \textbf{optimality rate} (OR) of a stable matching as its similarity to the proposer-optimal stable solution, which is unique.

\section{Generating Stable Solutions} \label{sec:generating}

The first task involves evaluating LLMs' abilities to generate valid, stable matchings in markets with various difficulty levels. This task ideally requires models to reason over ranked preferences while iteratively executing a structured algorithm.

We consider two sub-categories for generating matchings depending on declarative knowledge about algorithms:
i) prompt without specifying any algorithm, and
ii) prompt with exact step-by-step instructions on how to execute the DA algorithm \cite{gale1962college} (see \cref{sec:problem} for details). Our ablation studies showed that the above prompting strategies did not result in qualitatively different outcomes, as all models were able to correctly identify the requirement for considering preferences, the DA algorithm, and its execution steps.\footnote{Furthermore, converting the \textit{stable-marriage} setting to a \textit{task-scheduling} setting \citep{Fish2025EconEvals}, where ``men'' and ``women'' are replaced by ``workers'' and ``tasks'' (respectively), does not have a significant impact on performance.} The detailed results are presented in \Cref{app:CoT}.

\begin{figure*}[t]
    \centering
    \includegraphics[width=1\linewidth]{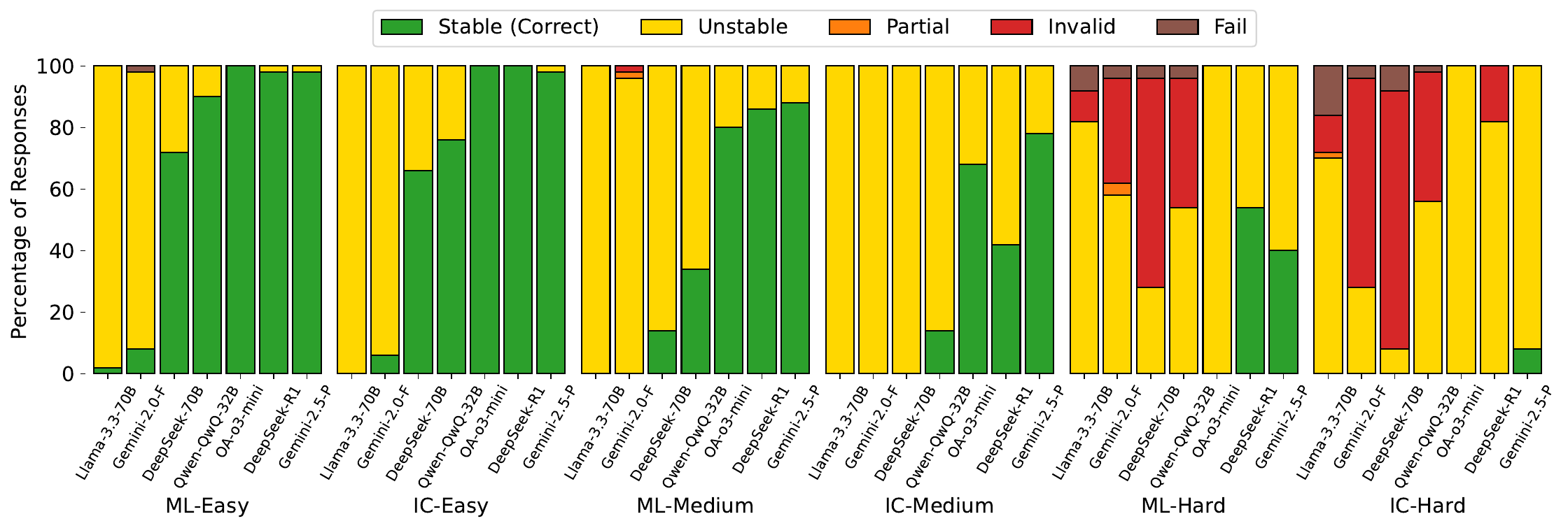}\caption{The generated responses by LLMs with Master-list (ML) and Impartial Culture (IC) preferences at different difficulty levels. 
    \textit{Stable} indicates one-to-one matchings with no blocking pairs; otherwise it is \textit{unstable}.
    \textit{Invalid} do not adhere to one-to-one constraint,  \textit{partial} are one-to-one but leave some unmatched, and \textit{Fail} indicates models' failure to return any matching.
    }
\label{fig:acheiving_bar}
\end{figure*}

\textbf{Difficulty, Model Size, and Reasoning.}
\Cref{fig:acheiving_bar} demonstrates the performance of the models in generating stable outcomes.
Baseline models without explicit reasoning mechanisms are unable to solve even Easy instances, whereas reasoning-enabled models achieve high accuracy on Easy instances but suffer dramatic performance drops on Hard instances. 
Furthermore, for Hard problems, even reasoning models frequently produce invalid outputs or fail to return any solution. Interestingly, \QwQ{} significantly outperforms a much larger model, \DSD, indicating that LLMs' combinatorial reasoning capability does not necessarily scale with model size.\footnote{Throughout the paper, all statistical comparisons between the percentages of stable solutions returned (across LLMs or across treatments) are done using Fisher's Exact test \citep{FishersExact}. Similarly, any two distributions of Instability or Optimality Rate are statistically compared using Welch's T-test \citep{WelchTTest}.}

\textbf{IC vs. ML Profiles.}
Under Impartial Culture (IC) profiles, the number of stable solutions can grow exponentially as the problem scales (increase of $n$) \cite{knuth1997stable}.
This combinatorial explosion poses a significant challenge for LLMs attempting to identify stable solutions, especially when solely using implicit reasoning over preference lists (without executing a concrete matching algorithm). 
In contrast, master‐list profiles (ML)---irrespective of the underlying sampling method used to generate preferences---admit exactly \textit{one} stable solution.
Moreover, this unique stable matching can be constructed in $\mathcal{O}(n)$ steps by (i) extracting the common Master-list and then (ii) pairing agents in order of their shared priority \citep{irving2008stable}. See the details of the algorithm in \Cref{app:algorithms}.

We observe that there is a significant performance gap between ML and IC instances---this disparity is especially marked for \Deepseek. 
With IC profiles and Hard instances, \textit{all} models are unable to compute a stable solution.\footnote{\GemPro{} is the only model with a positive success rate ($=8\%$) with IC preferences Hard instances.} They perform slightly better under ML profiles, and while this performance drops for Hard instances, these models almost never return invalid or partial matchings.

\textbf{Prompting Techniques.} 
Prompt-engineering techniques have been empirically demonstrated to enhance the performance of LLMs on mathematical reasoning and formal logic inference tasks \citep{wei2022chain,Yao2023ToT,Besta2024GoT}.
We evaluated a range of prompt-engineering techniques—including few-shot prompting and Chain-of-Thought (CoT) prompting, which supply exemplar ``thought processes'' and intermediate reasoning steps—in an attempt to bolster LLM performance. However, none of these strategies qualitatively improved on medium- or hard-difficulty instances. See \Cref{app:CoT} for further details.

\subsection{Measuring Instability}

\begin{figure}[t]
  \centering
  \begin{minipage}[t]{0.49\textwidth}
    \centering
    \includegraphics[width=\textwidth]{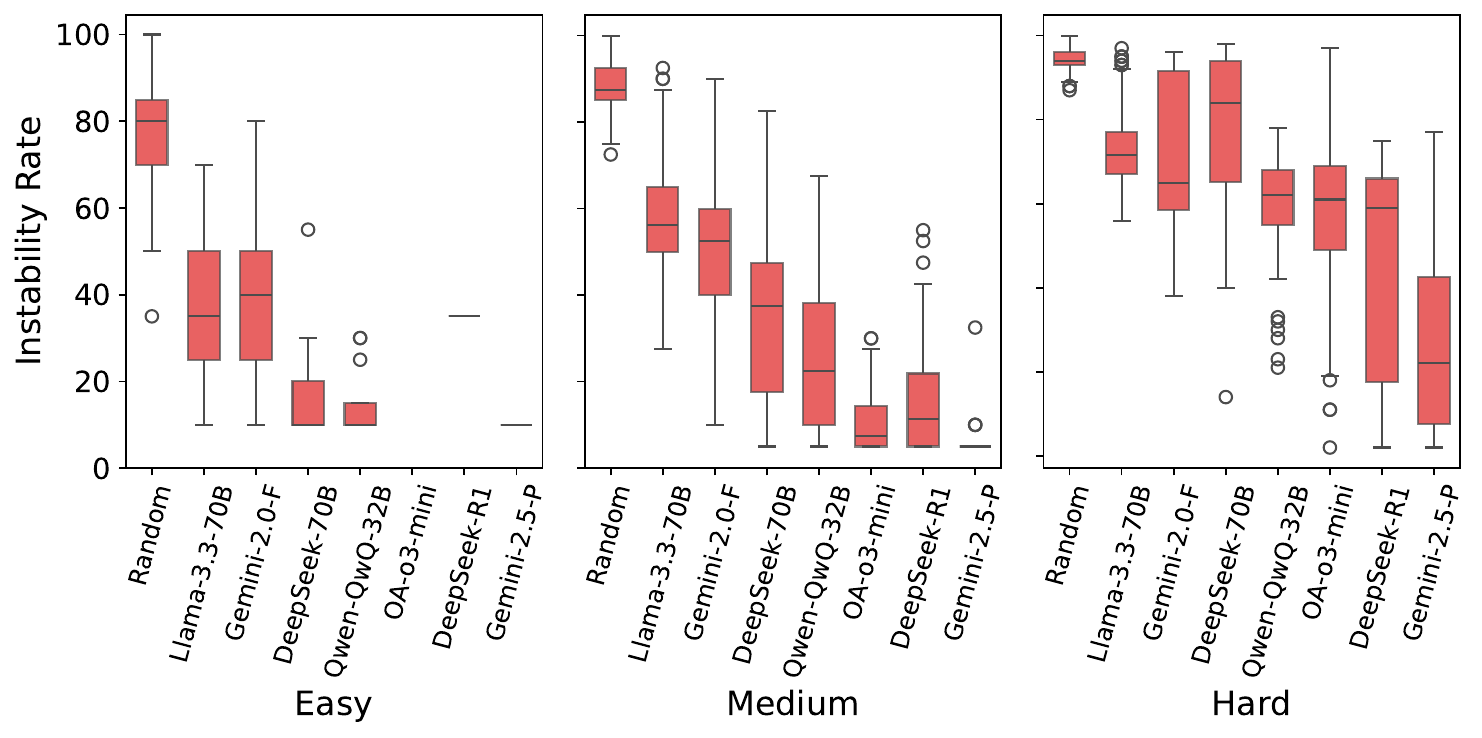}\caption{Instability Rate (lower is better) within unstable outcomes returned by each model as compared to randomly selected valid (but not necessarily stable) solutions (Random). 
    }
    \label{fig:bps}
  \end{minipage}
  \hfill
  \begin{minipage}[t]{0.49\textwidth}
    \centering
    \includegraphics[width=\textwidth]{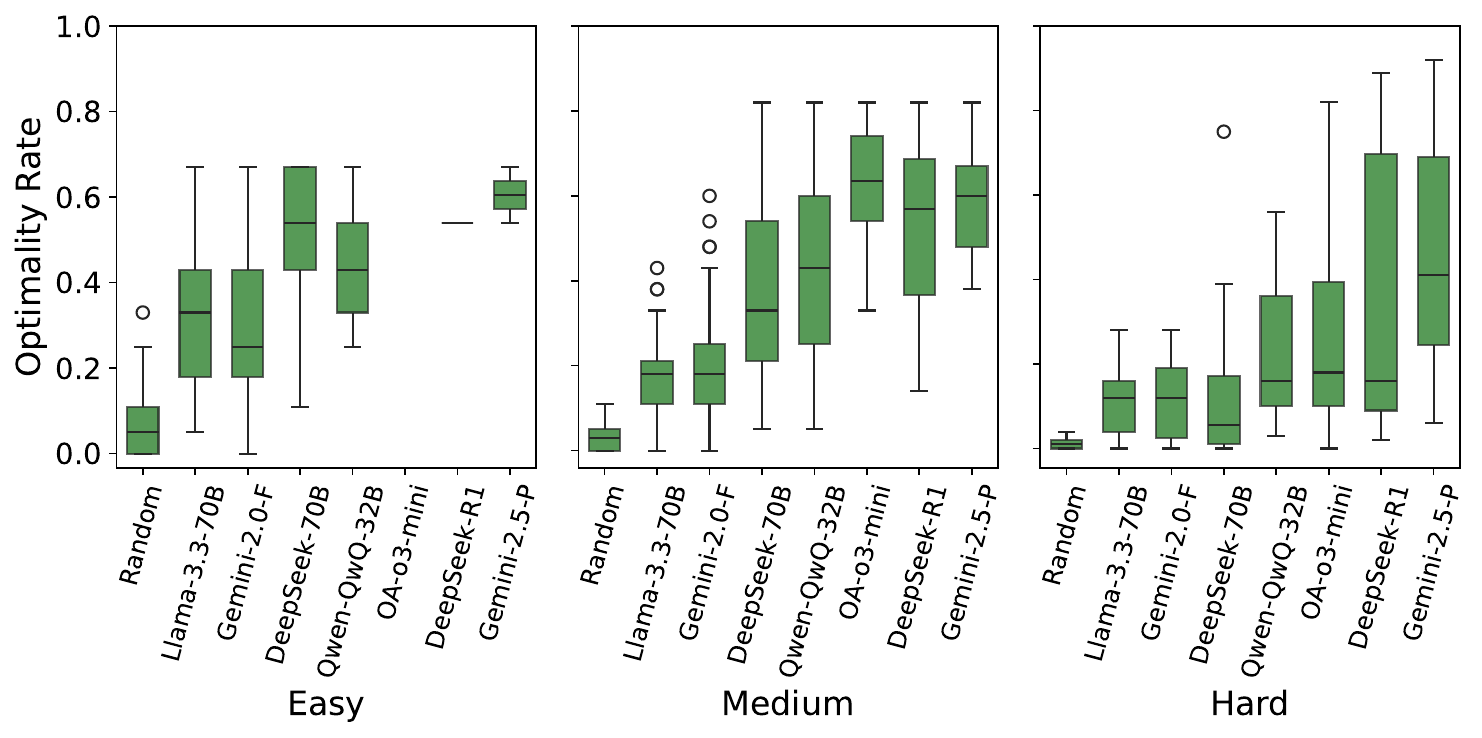}\caption{Optimality Rate within unstable outcomes returned by each model as compared to randomly selected valid (but not necessarily stable) solutions (Random).
    }
    \label{fig:intersections}
  \end{minipage}
\end{figure}

A natural question is how far LLM‐produced responses deviate from stable outcomes. To quantify this, we use two complementary metrics: instability rate and optimality rates (see \Cref{sec:profiles}). 
The \textbf{instability rate} directly reflects the distance from any stable solution, whereas the \textbf{optimality rate} implicitly evaluates the model's success in executing the underlying matching procedure.
\Cref{fig:bps} and \Cref{fig:intersections} illustrate comparisons of LLMs with a baseline of randomly generated outcomes.\footnote{Note that the plots only illustrate unstable but valid one-to-one outputs.}

Broadly, the advanced-reasoning models generate significantly closer approximations to stability and optimality than their non-reasoning counterparts.
Moreover, all evaluated LLMs (regardless of reasoning sophistication) substantially outperform random baselines on both metrics, indicating that they inherently leverage preference structures and exhibit nontrivial reasoning about ranked inputs.

Interestingly, the performance distinction between basic and reasoning models becomes less clear. While the intermediate reasoning models return a lower instability rate in Easy and Medium problems, their performance significantly drops in larger-scale problems (Hard). In fact, the performance of \DSD{} becomes worse than even basic non-reasoning models. We attribute this behavior to the model's diminished capacity for handling larger input lengths, a hypothesis supported by their lower proportion of valid outcomes (as seen in \Cref{fig:acheiving_bar}).

\section{Resolving Instability}\label{sec:resolving}

Generating stable solutions requires both exact reasoning over agents’ preference lists and the execution of a stability-guaranteeing procedure (e.g., the DA algorithm). As demonstrated in \Cref{sec:generating}, all evaluated models---irrespective of their reasoning capabilities---exhibit severe performance degradation as the problem size grows. 
This leads to the natural question of whether these models can resolve instability in a given matching---a task that entails detecting blocking pairs through preference reasoning and applying an appropriate sequence of adjustments to restore stability. 

We provide LLMs with unstable (but valid) matchings along with preference profiles, and instruct them to convert these initialized solutions to stable matchings. 
To assess how the instability rate may influence solution quality, we distinguish two classes of initial matchings: (i)
\textbf{One-BP}, matchings containing exactly one blocking pair (i.e., ``almost stable'') such that their stability may be resolved through simpler proposal-rejection iterations, and (ii)
\textbf{Random}, matchings sampled uniformly at random from the set of all valid one‐to‐one pairings, which typically contain a large number of blocking pairs and thus exhibit high degrees of instability. See \Cref{app:resolving} for detailed steps and pseudo-code for generating one-BP and random initialization.
Note that starting from an arbitrary matching, sequentially resolving blocking pairs may result in a cycle---as shown by \citet{knuth1976marriages}. However, a random sequence converges to stability with probability one \citep{roth1990random,abeledo1995paths}. 

\begin{table*}[t]
    \centering
    
    \caption{The percentage of stable matchings returned when tasked with resolving instability starting from One-BP or Random matchings. The numbers in bold represent the highest accuracy (across all LLMs) of resolving the corresponding type of unstable matching. 
    }
    \resizebox{\textwidth}{!}{
    \begin{tabular}{ll*{20}{c}}
        \toprule
        & & \multicolumn{5}{c}{\textbf{Basic LLMs}} & & \multicolumn{5}{c}{\textbf{Reasoning LLMs}} & & \multicolumn{8}{c}{\textbf{Advanced Reasoning LLMs}} \\\cmidrule{3-7}\cmidrule{9-13}\cmidrule{15-22}
        & & \multicolumn{2}{c}{\textbf{\GemFlash}} &
        & \multicolumn{2}{c}{\textbf{\Llama}} & 
        & \multicolumn{2}{c}{\textbf{\QwQ}} & 
        & \multicolumn{2}{c}{\textbf{\DSD}} &
        & \multicolumn{2}{c}{\textbf{\othree}} &
        & \multicolumn{2}{c}{\textbf{\Deepseek}} &
        & \multicolumn{2}{c}{\textbf{\GemPro}}  \\\cmidrule{3-4}\cmidrule{6-7}\cmidrule{9-10}\cmidrule{12-13}\cmidrule{15-16}\cmidrule{18-19}\cmidrule{21-22}
        \textbf{Difficulty} & \textbf{Preference} 
        & \textbf{One-BP} & \textbf{Random} &
        & \textbf{One-BP} & \textbf{Random} &
        & \textbf{One-BP} & \textbf{Random} &
        & \textbf{One-BP} & \textbf{Random} &
        & \textbf{One-BP} & \textbf{Random} &
        & \textbf{One-BP} & \textbf{Random} &
        & \textbf{One-BP} & \textbf{Random} \\
        \midrule
        \multirow{2}{*}{\textbf{Easy}} & \textbf{IC} & 2 & 2 & & 2 & 0 & & 60 & 36 & & 46 & 54 & & \textbf{100} & \textbf{100} & & 96 & 98 & & 96 & 92 \\
        & \textbf{ML} & 4 & 2 & & 0 & 0 & & 88 &78 & & 68 & 62 & & 96 & \textbf{100} & & \textbf{100} & 98 & & \textbf{100} & 98 \\\midrule
        \multirow{2}{*}{\textbf{Medium}} & \textbf{IC} & 0 & 0 & & 0 & 0 & & 22 & 0 & & 10 & 0 & & 64 & \textbf{64} & & 28 & 32 & & \textbf{74} & 60 \\
        & \textbf{ML} & 0 & 0 & & 0 & 0 & & 17 & 7 & & 20 & 6 & & 82 & 78 & & \textbf{88} & 76 & & 80 & \textbf{82} \\\midrule
        \multirow{2}{*}{\textbf{Hard}} & \textbf{IC} & 0 & 0 & & 0 & 0 & & \textbf{4} & 0 & & 0 & 0 & & 0 & 0 & & 0 & 0 & & 2 & \textbf{2} \\
        & \textbf{ML} & 0 & 0 & & 0 & 0 & & 0 & 0 & & 0 & 0 & & 6 & 0 & & \textbf{28} & 24 & & 16 & \textbf{34} \\\midrule
        \multicolumn{2}{c}{\textbf{Average}} & 1.00 & 0.67 & & 0.33 & 0.00 & & 31.83 & 20.16 & & 24.00 & 20.33 & & 58.00 & 57.00 & & 56.67 & 54.67 & & \textbf{61.33} & \textbf{61.33}  \\
        \bottomrule
        \end{tabular}
        }
    \label{tab:resolve_acc}
\end{table*}

\Cref{tab:resolve_acc} displays the fraction of responses in which LLMs return stable matchings when asked to resolve the above types of unstable matching.
Surprisingly, our experiments illustrate that in the task of resolving instability, the performance of all evaluated models does not exceed---and even degrades---their performance in generating stable solutions. This behavior persists regardless of initial matchings (One-BP or Random) and LLMs' reasoning capability.
In fact, on Hard instances, the output returned by advanced reasoning models on One-BP matchings (i.e., containing a single blocking pair) contains substantially more than one blocking pair.
In other words, even for the most basic instances, 
LLMs often introduce additional instabilities beyond the original single violation. We elaborate on this in \Cref{app:resolving}, demonstrating how all models, including those with advanced reasoning, often return solutions with a higher instability rate, highlighting their inability to systematically eliminate blocking pairs in accordance with preference lists.

\section{Detecting Instability}\label{sec:detecting}

The findings in previous sections raise the question of whether LLMs can reliably detect instability in a given matching---a simpler task that involves iterating over each unmatched pair to determine whether both agents prefer one another over their assigned partners.
This procedure entails only a straightforward comparison of preferences and requires $\mathcal{O}(n^2)$ steps.

\begin{figure}[h!]
    \centering
    \includegraphics[width=\linewidth]{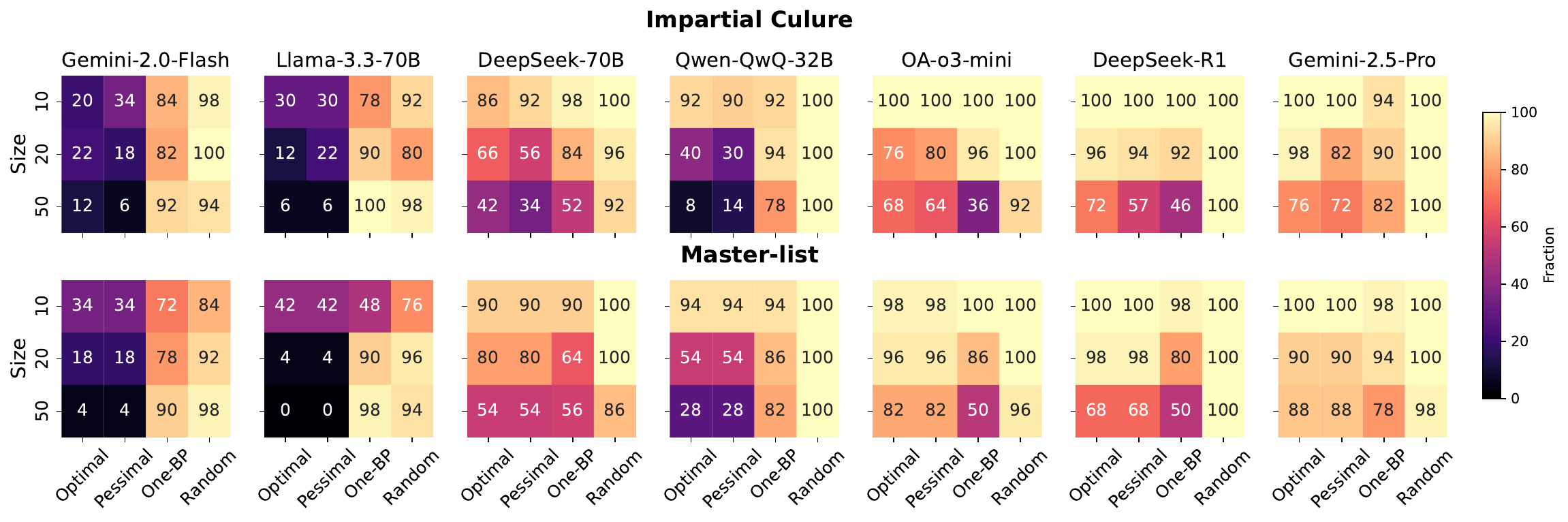}\caption[]{The fraction of responses where each model correctly detects stability or instability of a given matching.}
    \label{fig:evaluating_heatmap}
\end{figure}

For this task, we evaluate the performance of valid one-to-one matchings initialized under two instability conditions: (i) One-BP, representing nearly stable matchings containing a single blocking pair, and (ii) Random, representing highly unstable matchings with numerous blocking pairs. 
To detect false-positives, we additionally include two extreme cases of stable matchings: the proposer-optimal (Optimal) and the proposer-pessimal (Pessimal) stable solutions.

\textbf{Reasoning Models: Preferences and Blocking Pairs.} 
\Cref{fig:evaluating_heatmap} reveals an interesting observation about reasoning models: their performance is influenced by the number of blocking pairs in the matching being evaluated---similar to the observations in \Cref{sec:resolving}.
They achieve a high accuracy on identifying random matchings (which have a larger number of blocking pairs) as unstable and a significantly lower accuracy with matchings that have at most one blocking pair.

\textbf{Basic Models and Hallucination.} Interestingly, the non-reasoning models achieve a high accuracy ($80\%$) with both types of unstable matchings, and extremely low ($20\%$) accuracy with stable matchings. 
Note that the performance across all models is qualitatively similar in ML and IC profiles, even though each ML profile admits a unique stable solution (thus, identical reports for Optimal and Pessimal). See \cref{app:detecting} for further analysis.
A manual analysis of non‐reasoning models uncovers frequent hallucinations about blocking pairs, resulting in a systematic bias toward classifying matchings as unstable. 
This can be largely attributed to misinterpretations of the input preferences.

\section{Reasoning about Ranked Preferences} \label{sec:reasoningPref}

Many advanced reasoning paradigms, ranging from causal inference \citep{Haoang2024Causal} and counterfactual analysis to game-theoretic decision making, depend fundamentally on the ability to compare and evaluate alternative choices.
As demonstrated thus far, even advanced reasoning models often fail to execute the step-by-step procedures of combinatorial algorithms when those procedures operate over ranked preference lists.
This shortcoming motivates the question of whether current LLMs can truly reason \textit{about} preferences, as opposed to merely applying preferences in generating responses heuristically. 

To investigate preference comprehension, we introduce a suite of tasks spanning three levels of inference over ranked preferences: 
(i) \textbf{basic retrieval} \textbf{(L1)}, in which models must extract individual preference relations; 
(ii) \textbf{comparison queries} \textbf{(L2)}, requiring pairwise preference judgments; and 
(iii) \textbf{proposal-acceptance simulations} \textbf{(L3)}, which combine comparison of alternatives with binary accept/reject decisions mirroring the dynamics of deferred-acceptance algorithms.

Hierarchical, level‐wise reasoning evaluations have been proposed recently in domains such as causal inference of LLMs  \citep{Haoang2024Causal}. For example, an \textbf{L1} question is ``\textit{Who is agent W5's, 4th-most preferred agent?}'', and an \textbf{L2} question ``Would agent W5, prefer M8 over M7?''

\begin{figure}[h!]
    \centering
    \includegraphics[width=\linewidth]{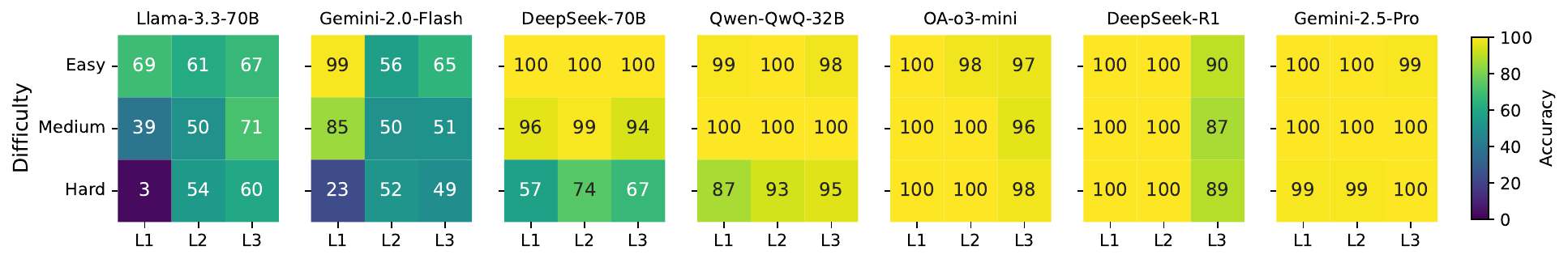}\caption{Accuracy on questions about provided preferences, with both IC and ML instances.}
    \label{fig:preference_heatmap}
\end{figure}

Basic models have low accuracy in all levels of difficulty even on small instances, which is probably the reason behind their inability to compute or detect stable solutions (as discussed in \Cref{sec:generating} and \Cref{sec:detecting}).
In basic models (e.g., \Llama) and (non-advanced) reasoning models (e.g., \DSD), the size of the problem has a greater impact on accuracy as compared to the level of the question, indicating their difficulty of handling larger inputs.
Although advanced-reasoning models have significantly higher accuracy rates compared to other models, they still make minor errors, especially in larger profiles (Hard). 
Generating stable solutions for these instances often requires a larger number of reasoning steps over many preference lists, causing minor errors to compound (small errors multiply!).

\section{Performance Improvement through LoRA Fine-Tuning} \label{sec:fine-tuning}

Supervised fine-tuning has proven to be effective in enhancing the reasoning capabilities of LLMs in a variety of tasks, including mathematical problem solving \citep{Chung2024Flan,Lewkowycz2022Quant}, logical reasoning \citep{Morishita2023Logical}, and code generation \citep{Luo2024Wizard}. \citet{Markeeva2024CLRS} demonstrate how fine-tuning a small LLM (2B parameters) can significantly improve LLMs' ability to execute textbook algorithms (e.g., sorting an array, finding the shortest path between two nodes on a graph, etc.). Hence, we evaluate whether fine-tuning can be used to enhance LLMs' ability to compute stable matchings---a task requiring reasoning over ranked preferences in addition to algorithmic understanding. To this end, we perform LoRA fine-tuning \citep{Hu2022LoRA} on three reasoning models, including \QwQ{} and two smaller models, \DSLlamaSmall{} and \DSQwenMed, for Generating Stable Solutions task. Additionally, we also evaluate whether fine-tuning can mitigate errors made by LLMs in the Preference Reasoning task (\Cref{sec:reasoningPref}).

\textbf{Training.}  Let $\mathcal D = \{(\mathbf x^{(i)},\mathbf y^{(i)})\}_{i=1}^N$ be the fine-tuning dataset for a given task.
Each pair consists of an \emph{input instance} $\mathbf x^{(i)}$ and the \emph{desired model completion} $\mathbf y^{(i)}$. The input instance $\mathbf x^{(i)}$ is made up of four components: (i) a generic \textit{system-prompt} $s$, (ii) a high-level \textit{instruction} $u$, (iii) the \textit{preference profile} $p^{(i)}$, and (iv) the \textit{task-prompt} $t^{(i)}$. The desired completion $\mathbf y^{(i)}$ consists of two components, (i) a chain-of-thought \textit{reasoning trace} $r^{(i)}$, and (ii) the \textit{answer} $a^{(i)}$ in the desired format. Each model is fine-tuned with standard next-token cross-entropy on the concatenated sequence $\mathbf z^{(i)} = \mathbf x^{(i)} || \mathbf y^{(i)}$. We separately fine-tune each model for the Generating Stable solutions task ($N = 10,000$) and the Preference Reasoning task ($N = 9,000$). See \Cref{app:ft_details} for details of the fine-tuning setup and results from experiments with different hyperparameter values.

\textbf{Improvement.} Fine-tuning LLMs with data containing synthetically generated reasoning traces substantially improves their performance on both tasks, as evidenced in \Cref{tab:ft_enhancement}. 
\begin{wraptable}{r}{0.6\textwidth}
\centering
\caption[]{Improvement in performance in the Generating Stable Solutions and Preference Reasoning tasks after fine-tuning on respective datasets.}
\resizebox{0.6\textwidth}{!}{
\begin{tabular}{ll*{9}{c}}
\toprule
\multirow{3}{*}{\textbf{Model}} & \multirow{3}{*}{\textbf{Stage}}
  & \multicolumn{6}{c}{\textbf{\makecell{Generating\\Stable Solutions}}}
  & \multicolumn{3}{c}{\textbf{\makecell{Preference\\Reasoning}}} \\
\cmidrule(lr){3-8}\cmidrule(lr){9-11}
 & & \multicolumn{3}{c}{\textbf{Stable Solutions (\%)}} 
   & \multicolumn{3}{c}{\textbf{Instability Rate ($\downarrow$)}} 
   & \multicolumn{3}{c}{\textbf{Accuracy (\%)}} \\
\cmidrule(lr){3-5}\cmidrule(lr){6-8}\cmidrule(lr){9-11}
 & & Easy & Med. & Hard & Easy & Med. & Hard & Easy & Med. & Hard \\
\midrule
\multirow{2}{*}{\textbf{\DSLlamaSmall}} 
 & \textbf{Vanilla}     & 3.0  & 0.0  & 0.0 & 41.02 & 64.19 & 92.70 & 81.67 & 74.33 & 47.67 \\
 & \textbf{Fine-tuned}  & 64.0 & 31.0 & 0.0 & 6.00 & 12.13 & --    & \textbf{100.0} & 98.33 & 75.00 \\
\midrule
\multirow{2}{*}{\textbf{\DSQwenMed}} 
 & \textbf{Vanilla}     & 19.0 & 0.0  & 0.0 & 20.66 & 55.31 & 94.09 & 97.67 & 91.33 & 72.33 \\
 & \textbf{Fine-tuned}  & 51.0 & 41.0 & 0.0 & 16.35 & 24.42 & 84.00 & \textbf{100.0} & \textbf{100.0} & 91.00 \\
\midrule
\multirow{2}{*}{\textbf{\QwQ}} 
 & \textbf{Vanilla}     & 83.0 & 24.0 & 0.0 &  2.35 & 19.27 & 59.07 &  99.00 & \textbf{100.0} & 91.67 \\
 & \textbf{Fine-tuned}  & \textbf{100.0}& \textbf{100.0}& 0.0 &  \textbf{0.00} &  \textbf{0.00} & \textbf{55.19} & \textbf{100.0} & \textbf{100.0} & \textbf{99.00} \\
\bottomrule
\end{tabular}
}
\label{tab:ft_enhancement}
\end{wraptable} In fact, this approach enhances the performance of \QwQ, a (non-advanced) reasoning model, to a success rate of $100\%$ in computing stable matchings with \textit{both} Easy and Medium instances, significantly outperforming advanced-reasoning models. Fine-tuning also clearly improves smaller models, i.e. \DSLlamaSmall{} and \DSQwenMed, both in terms of achieving stability and Instability Rate.\footnote{Interestingly, this improvement is clearer for ML instances, where both models achieve a $100\%$ success rate at the Easy level and $>80\%$ success rate at the Medium level.} Similar results are obtained for the Preference Reasoning tasks, with the error-rate reducing to $0$ at the Easy and Medium levels.
\footnote{The only exception being \DSLlamaSmall{} at the Medium level.}

In spite of these improvements, however, there remains a distinct gap in performance at the Easy and Medium levels as compared to the Hard level. LLMs remain altogether unable to compute stable matchings for Hard instances, even after fine-tuning. A similar trend is reflected in the accuracy on the Preference Reasoning task. Hence, while fine-tuning clearly improves the reasoning capabilities of LLMs, further enhancements are required to improve their ability to handle larger inputs.

\begin{table*}[t]
    \centering
    
    \caption{The performance of LLMs with different reasoning capability across all tasks requiring reasoning over ranked preferences and executing structured algorithms. 
    }
    \resizebox{\textwidth}{!}{
    \begin{tabular}{llccccccccccc}
        \toprule
        & & \multicolumn{3}{c}{\textbf{\makecell{Generating\\Stable Solutions}}} & &\multicolumn{3}{c}{\textbf{\makecell{Resolving\\Instability}}} &  &\textbf{\makecell{Detecting\\Instability}} & & \textbf{\makecell{Preference\\Reasoning}} \\
        \cmidrule{3-5}\cmidrule{7-9}\cmidrule{11-11}\cmidrule{13-13}
        \textbf{Category} & \textbf{Model} & \textbf{\makecell{Stable\\Solutions ($\%$)}} & \textbf{\makecell{Instability\\Rate ($\downarrow$)}} & \textbf{\makecell{Optimality\\Rate ($\uparrow$)}} & &\textbf{\makecell{Stable\\Solutions ($\%$)}} & \textbf{\makecell{Instability\\Rate ($\downarrow$)}} & \textbf{\makecell{Optimality\\Rate ($\uparrow$)}} & & \textbf{Accuracy ($\%$)} & & \textbf{Accuracy ($\%$)} \\
        \midrule
        \multirow{2}{*}{\textbf{Basic}}& \textbf{Llama-3.3-70B}    &  0.33 &  54.03 &   0.21 & & 0.17 &  56.04 &   0.25 & & 56.76 & & 52.67 \\
        & \textbf{Gemini-2.0-Flash} &  2.36 &  48.44 & 0.21 & & 0.83 &   53.94 &   0.22 & & 58.38 & & 58.89 \\\midrule
        \multirow{2}{*}{\textbf{Reasoning}} &  \textbf{DeepSeek-70B} & 26.20 & 23.43 & 0.59 & & 22.49 & 24.61 & 0.60 & & 77.05 & & 88.67  \\
        & \textbf{Qwen-QwQ-32B}  & 35.67 &  21.22 & 0.63 & & 28.00 & 26.37 & 0.62 & & 75.05 & & 96.89  \\\midrule
        \multirow{3}{*}{\textbf{\makecell{Advanced\\Reasoning}}} &\textbf{o3-mini} & 58.00 & 19.98 & 0.72 & & 57.50 & 18.52 & 0.75 & & 86.67 & & 98.78 \\
        & \textbf{DeepSeek-R1} & 64.22 & 12.35 & 0.80 & & 55.73 & 14.21 & \textbf{0.79} & & 88.19 & & 96.22 \\
        & \textbf{Gemini-2.5-Pro}   & \textbf{68.33} &    \textbf{7.16} &  \textbf{0.84} & & \textbf{61.33} &   \textbf{8.80} &  \textbf{0.79} & & \textbf{92.38} & & \textbf{99.67} \\
        \bottomrule
    \end{tabular}}
    \label{tab:overall}
\end{table*}

\section{Concluding Remarks}

We summarize the performance of LLMs across all four tasks in \Cref{tab:overall}, reflecting the clear hierarchy between advanced-reasoning models, (non-advanced) reasoning models, and basic models.
The limitations in reliably reasoning about ranked preferences raise concerns about the viability of LLMs as agents acting on behalf of users in market-oriented or preference-sensitive decision-making settings, limit their capacity to negotiate complex user preferences, and hinder efforts in developing pluralistic techniques (e.g., constitutional AI \citep{bai2022constitutional} and social choice-theoretic \citep{Conitzer+24}) for value alignment that are inherently based on aggregating rankings. 

\textbf{Open-Source vs. Closed-Source models.} Among the models that we evaluate, \GemPro{} (a closed-source model) emerges as the most capable across all tasks. While \Deepseek{} (open-source) broadly outperforms OpenAI's \othree{} (closed-source), it performs much worse with IC instances than with ML instances. While both basic models struggle on all tasks, \GemFlash{} (closed-source) marginally outperforms \Llama{} (open-source) on various metrics. Given the promising improvement yielded by fine-tuning an open-source reasoning model, i.e. \QwQ, it is worth exploring strategies that also enable it to handle large inputs.

\textbf{Beyond Linear Preferences.} Our current evaluation paradigm considers complete and strict linear preferences. In real-world scenarios, however, preferences involve complexities such as incompleteness, indifference between alternatives, and capacity constraints \citep{Delorme2019Incomplete,Kawanashie2013Hospital,Manlove2O22Student}. As a result, algorithms to compute stable solutions in such settings are far more complex and solutions are often intractable \citep{Manlove2014Book}. While, in \Cref{app:ties}, we provide some preliminary insights on preferences with ties, understanding such cases requires deeper theoretical and empirical investigation. A meaningful next step would be to examine how AI models respond to these more intricate preference structures.

\section*{Acknowledgments}
This research was supported by the National Science Foundation (NSF) through CAREER Award IIS-2144413 and Award IIS-2107173.
We thank the anonymous reviewers for their fruitful comments. We would also like to thank Shraddha Pathak for several useful discussions.

\bibliographystyle{plainnat}
\bibliography{main}

\newpage
\appendix

\section{Limitations and Future Work}

While we present a comprehensive evaluation of the practical algorithmic and economic reasoning capabilities of a series of state-of-the-art LLMs, our dataset primarily relies on synthetic data due to the challenges in obtaining real-world ordinal preference data. This calls for the collection and curation of datasets in the two-sided matching setting, and generating preference profile datasets that are better aligned with human preferences. 

Additionally, while our work provides insights into the reasons behind the failure of LLMs to consistently generate stable solutions (see \Cref{sec:reasoningPref}), there is scope for further clarity on where exactly LLMs make mistakes during algorithmic execution. A potential method to explore this is to break the algorithmic execution task into smaller steps (e.g., a single proposal-acceptance/rejection cycle) and identify which components of the \textit{state-transition} are challenging for LLMs to understand. 

Furthermore, while fine-tuning substantially enhances LLMs' performance on instances with relatively smaller input sizes, improving their performance with larger inputs requires further exploration. This can include the investigation of methods such as fine-tuning the entire set of parameters (unlike with LoRA) or reinforcement-learning methods such as \textit{group-relative policy optimization} (GRPO) that are known to increase the reasoning capabilities of LLMs \citep{shao2024deepseekmathpushinglimitsmathematical}.

\section{Broader Impacts}
This paper is intended to advance Machine Learning and AI research, with a special emphasis on the reasoning ability of LLMs---an essential component of autonomous AI systems. We identify key shortcomings in the reasoning capabilities of LLMs, especially in terms of aggregating individuals preferences over alternatives and algorithmic execution. We believe that the findings presented in this work can inform further research into AI systems to enhance their ability to act independently in complex decision-making environments.

\section{Preference-Based Algorithms in Matching Markets} \label{app:algorithms}

\subsection{Algorithm for Generating a Stable Matching}
As shown in \Cref{alg:deferred-acceptance}, the standard deferred-acceptance algorithm runs by having the proposing side of the market make a series of proposals, and each agent that receives at least one proposal decides which proposal to accept (the proposal becomes an engagement), and which proposals to reject (or engagements to break). This continues until all agents are matched, which requires $O(n^2)$ proposal steps. The resulting solution is stable \cite{gale1962college}. To describe the algorithm, we adopt the traditionally used setting of \textit{stable-marriage} where the proposing side consists of \textit{men} and the receiving side consist of \textit{women}.

\begin{algorithm}
\caption{The Deferred Acceptance Algorithm}\label{alg:deferred-acceptance}
\begin{algorithmic}
\State{assign each agent $m \in M$ and $w \in W$ to be free}
\While{there exists a free man $m$ who has not proposed to every woman}
    \State $w \leftarrow$ highest-ranked woman on $m$'s preference list to whom he has not yet proposed
    \State $m$ proposes to $w$
    \If{$w$ is free}
        \State $w$ tentatively accepts $m$
    \ElsIf{$w$ prefers $m$ to her current partner $m'$}
        \State $w$ rejects $m'$ and tentatively accepts $m$
        \State $m'$ becomes free
    \Else
        \State $w$ rejects $m$
    \EndIf
\EndWhile
\State \textbf{return} the set of engaged pairs, these form a stable matching
\end{algorithmic}
\end{algorithm}

Another commonly used version of the DA algorithm involves the receiver agent (in a given proposal) removing all agents (on the proposer-side) from their preference list who are ranked below the current proposer agent (who also remove the receiver from their respective preference lists). Due to the shortening of the preference lists as the algorithm progresses (see \Cref{alg:deferred-acceptance-sl}), this is referred to as the version of the DA algorithm \textit{with Shortlists} \citep{ESM}. While this version terminates with at most as many (and often less) proposal steps as compared to \Cref{alg:deferred-acceptance}, it requires repeated updates to the original preference lists.  

\begin{algorithm}
\caption{The Deferred-Acceptance Algorithm with Shortlists}\label{alg:deferred-acceptance-sl}
\begin{algorithmic}
\State{assign each agent $m \in M$ and $w \in W$ to be free}
\While{some man $m$ is free}
    \State{$w$ = first woman on $m$'s preference list}
    \State{$m$ proposes and becomes engaged to $w$}
    \If{some man $p$ is engaged to $w$}
        \State{break the engagement $(p,w)$, assign $p$ to be free}
    \EndIf
    \For{each $m'$ in $w$'s preference list s.t. $m 
    \succ_w m' $}
        \State{remove $m'$ and $w$ from each other's preferences}
    \EndFor
\EndWhile
\State \textbf{return} the set of engaged pairs, these form a stable matching
\end{algorithmic}
\end{algorithm}

\subsection{Algorithm for Generating a Stable Matching w/ Master-Lists}
Similar to \Cref{alg:deferred-acceptance}, \Cref{alg:da-master} runs in rounds of proposals. Since there is a Master-list over proposing agents in the preference lists of receiving agents, and proposing agents are selected to make proposals in the order in which they appear in the Master-list, there are no rejections (since any receiver receives the best possible proposal at any step) \citep{irving2008stable}. Hence, this algorithm terminates in $n$ proposal steps (and is therefore easier to execute).

\begin{algorithm}
\caption{The Deferred-Acceptance Algorithm for Preferences w/ Master-lists on One Side}\label{alg:da-master}
\begin{algorithmic}
\State{assign each agent $m \in M$ and $w \in W$ to be free}
\State{$L \gets$ Master-list over men.}
\For{next free man $m$ in $L$}
    \State{$w$ = first woman on $m$'s preference list}
    \State{$m$ proposes and becomes engaged to $w$}
\EndFor
\State \textbf{return} the set of engaged pairs, these form a stable matching
\end{algorithmic}
\end{algorithm}

\subsection{Algorithm for Resolving Instability}
While we don't explicitly describe the algorithm here, the mechanism presented by \citet{abeledo1995paths} can be applied to an unstable matching $\mu$ by resolving blocking pairs, resulting in a stable solution. Intuitively, an LLM does not have to follow a specific mechanism, rather the model can resolve instability by iteratively resolving blocking pairs as they arise (eventually, assuming all steps are correct, the model should arrive at a stable solution).

\subsection{Algorithm for Detecting Instability}
Intuitively, \Cref{alg:detect-stable} works by iteratively visiting each pair of agents $(m,w)$ s.t. $m \in M$ and $w \in W$, and finding a pair such that either $m$ prefers $w$ to their current partner, or $w$ prefers $m$ to their current partner (when such a pair is found, output it as a blocking pair). If no pair $(m,w)$ is found to be a blocking pair, then the solution is stable.

\begin{algorithm}
\caption{Stability Detecting Algorithm}\label{alg:detect-stable}
\begin{algorithmic}
\State{for each $(m,w) \in \mu$, where $m \in M$ and $w \in W$}
\For{man $m \in M$}
    \For{man $w \in W$}
        \If{$m \succ_w \mu(w)$ and $w \succ_m \mu(m)$}
            \State{output the identified blocking pair $(m, w)$}
        \EndIf
    \EndFor
\EndFor
\State{output that there exist no blocking pairs}
\end{algorithmic}
\end{algorithm}

\section{Prompt Engineering}\label{app:CoT}

\paragraph{Providing Algorithmic Description.} In \Cref{tab:with_algo}, examine LLMs' performance when provided with a prompt containing pseudocode for the DA algorithm, and compare it to the case when no algorithm is provided in the prompt. While providing the DA algorithm as part of the prompt leads to an improvement in some cases, the only case in which there is improvement is statistically significant\footnote{At $p < 0.05$, using Fisher's exact test} is with \DSD{} at the Easy level with ML instances.

\begin{table*}[t]
    \centering
    
    \caption{Percentage of stable solutions returned by LLMs when provided with the DA algorithm in the prompt (With) as compared to the case when not provided (Without). 
    }
    \resizebox{\textwidth}{!}{
    \begin{tabular}{cc*{20}{c}}
        \toprule
        & & \multicolumn{5}{c}{\textbf{Basic LLMs}} & & \multicolumn{5}{c}{\textbf{Reasoning LLMs}} & & \multicolumn{8}{c}{\textbf{Advanced Reasoning LLMs}} \\\cmidrule{3-7}\cmidrule{9-13}\cmidrule{15-22}
        & & \multicolumn{2}{c}{\textbf{\GemFlash}} &
        & \multicolumn{2}{c}{\textbf{\Llama}} & 
        & \multicolumn{2}{c}{\textbf{\QwQ}} & 
        & \multicolumn{2}{c}{\textbf{\DSD}} &
        & \multicolumn{2}{c}{\textbf{\othree}} &
        & \multicolumn{2}{c}{\textbf{\Deepseek}} &
        & \multicolumn{2}{c}{\textbf{\GemPro}}  \\\cmidrule{3-4}\cmidrule{6-7}\cmidrule{9-10}\cmidrule{12-13}\cmidrule{15-16}\cmidrule{18-19}\cmidrule{21-22}
        \textbf{Difficulty} & \textbf{Preference} 
        & \textbf{Without} & \textbf{With} &
        & \textbf{Without} & \textbf{With} &
        & \textbf{Without} & \textbf{With} &
        & \textbf{Without} & \textbf{With} &
        & \textbf{Without} & \textbf{With} &
        & \textbf{Without} & \textbf{With} &
        & \textbf{Without} & \textbf{With} \\
        \midrule
        \multirow{2}{*}{\textbf{Easy}} & \textbf{IC} & 6 & \textbf{10} & & \textbf{2} & 0 & & 76 & \textbf{84} & & 70 & \textbf{74} & & \textbf{100} & 98 & & \textbf{100} & 96 & & 98 & \textbf{100} \\
        & \textbf{ML} & \textbf{8} & 6 & & 2 & \textbf{6} & & \textbf{90} & 88 & & 72 & \textbf{94} & & 96 & \textbf{100} & & 98 & \textbf{100} & & 98 & 98 \\\midrule
        \multirow{2}{*}{\textbf{Medium}} & \textbf{IC} & 0 & 0 & & 0 & 0 & &\textbf{14} &2 & & 0 & 0 & & \textbf{68} & 64 & & 42 & \textbf{44} & & \textbf{90} & 88 \\
        & \textbf{ML} & 0 & 0 & & 0 & 0 & &34 &\textbf{40} & & 14 & 12 & & 80 & \textbf{86} & & \textbf{86} & 82 & & 88 & \textbf{94} \\\midrule
        \multirow{2}{*}{\textbf{Hard}} & \textbf{IC} & 0 & 0 & & 0 & 0 & & 0 & 0 & & 0 & 0 & & 0 & 0 & & 0 & 0 & & \textbf{8} & 6 \\
        & \textbf{ML} & 0 & 0 & & 0 & 0 & & 0 & 0 & & 0 & 0 & & 0 & 0 & & \textbf{54} & 36 & & \textbf{40} & 38 \\\midrule
        \multicolumn{2}{c}{\textbf{Average}} & 2.33 & \textbf{2.67} & & 0.67 & \textbf{1.00} & & 35.67 & 35.67 & & 26.00 & \textbf{30.00} & & 57.33 & \textbf{58.00} & & \textbf{63.33} & 59.66 & & 68.33 & \textbf{70.66}  \\
        \bottomrule
        \end{tabular}
        }
    \label{tab:with_algo}
\end{table*}

\paragraph{Reasoning-enhancement Prompts.} For models that fail at generating stable matchings even with small instances, we evaluate whether prompt-based enhancements such as Chain-of-Thought (CoT) \citep{wei2022chain} and Few-shot prompting \citep{dong2022survey} can improve their performance. In particular, we introduce the following three types of modifications to the original prompt (used in \Cref{sec:generating}):
\begin{itemize}
    \item \textbf{CoT-Vanilla (CoT-V):} The steps of execution of the DA algorithm (see \Cref{alg:deferred-acceptance}) are provided for an example instance. Each step consists of a single (free) proposing agent making his next proposal, and the receiving agent either accepting or rejecting the proposal based on their current status. 
    \item  \textbf{CoT-Shortlist (CoT-SL):} This version of CoT uses the Shortlist version of the DA algorithm (see \Cref{alg:deferred-acceptance-sl}) which reduces the overall number of proposal steps, but requires repeated updates to the original preference lists.
    \item \textbf{Few-shot Examples:} As opposed to the previous two cases, we provide LLMs with three examples of stable matching instances accompanied by their solutions.
\end{itemize}

To limit the context size of the prompt, we consider examples with $n=5$ for each of these  prompt modifications.

\begin{table*}
\centering
\scriptsize
\caption[]{Percentage of stable solutions returned when prompt-enhancement strategies are used, as compared to the case without, for the Easy and Medium difficulty levels. \footnotemark}
\resizebox{\textwidth}{!}{
\begin{tabular}{cccccccccccccccccccccc}\toprule
\textbf{} &\textbf{Model} & &\multicolumn{4}{c}{\textbf{\GemFlash}} & &\multicolumn{4}{c}{\textbf{\Llama}} & &\multicolumn{4}{c}{\textbf{\QwQ}} & &\multicolumn{4}{c}{\textbf{\DSD}} \\\cmidrule{2-2}\cmidrule{4-7}\cmidrule{9-12}\cmidrule{14-17}\cmidrule{19-22}
\textbf{Size} &\textbf{Culure} & &\textbf{None} &\textbf{CoT-V} &\textbf{CoT-SL} &\textbf{Few-shot} & &\textbf{None} &\textbf{CoT-V} &\textbf{CoT-SL} &\textbf{Few-shot} & &\textbf{None} &\textbf{CoT-V} &\textbf{CoT-SL} &\textbf{Few-shot} & &\textbf{None} &\textbf{CoT-V} &\textbf{CoT-SL} &\textbf{Few-shot} \\\midrule
\multirow{2}{*}{\textbf{10}} &\textbf{IC} & &\textbf{6} &2 &0 &0 & &0 &\textbf{2} &0 &0 & &\textbf{76} & 86 & 84 & \textbf{90} & &\textbf{70} & 60 & 60 & 68 \\
&\textbf{ML} & &\textbf{8} & 4 & 4 & 2 & & 2 &0 &\textbf{10} & 6 & & 90 &\textbf{94} & 92 & \textbf{94} & & 72 &\textbf{76} & 68 &\textbf{76} \\\midrule
\multirow{2}{*}{\textbf{20}} &\textbf{IC} & &0 &0 &0 &0 & &0 &0 &0 &0 & &\textbf{14} &2 &6 &6 & &0 &0 &0 &0 \\
&\textbf{ML} & &0 &0 &0 &0 & &0 &0 &0 &0 & &34 &36 &34 &\textbf{42} & &\textbf{14} &10 &4 &2 \\
\bottomrule
\end{tabular}}
\label{tab:cot-fs}
\end{table*}

\footnotetext{All models considered here are never able to generate stable matchings at the Hard difficulty level, with any prompting method.}

As depicted in \Cref{tab:cot-fs}, these prompting enhancements fail to improve the ability of LLMs to generate stable matchings. While there is a slight improvement for models like \QwQ{} and \DSD{} instances with Master-list preferences and size $n=10$, this improvement is not statistically significant.  

\paragraph{Modified Problem Setting.} We consider the traditionally used setting of \textit{stable-marriage}, where the set $M$ consists of \textit{men} who propose to \textit{women} in the set $W$ \citep{GS62college}. To measure whether LLMs are sensitive to the nomenclature used to described the two-sided matching market, we also consider a different setting, i.e. that where a set of \textit{workers} ($W$) are to be assigned a set of \textit{tasks} ($T$) (and members on both sides have preferences over members of the other). We test the difference in the performance of two LLMs---\GemFlash{} and \othree---between the task-scheduling and stable-marriage settings. The results are provided in \Cref{tab:workermen}. While there is a slight decrease in performance for \othree, the change is not significant (at $p < 0.05$). This provides further evidence that LLMs understand requirements of computing stable solutions for matching markets, in general.

\begin{table*}
\centering
\scriptsize
\caption{Percentage of stable solutions from two LLMs when the task is framed as the \textit{stable marriage} and the \textit{task-scheduling problem}.}

\label{tab:workermen}
\resizebox{0.5\textwidth}{!}{
\begin{tabular}{lcccccccc}\toprule
\textbf{} &\textbf{Model} & &\multicolumn{2}{c}{\textbf{Gemini-2.0-F}} & &\multicolumn{2}{c}{\textbf{o3-mini}} \\\cmidrule{2-2}\cmidrule{4-5}\cmidrule{7-8}
\textbf{Difficulty} &\textbf{Preference} & &\textbf{\makecell{Stable\\Marriage}} &\textbf{\makecell{Task\\Scheduling}} & &\textbf{\makecell{Stable\\Marriage}} &\textbf{\makecell{Task\\Scheduling}} \\\midrule
\multirow{2}{*}{\textbf{10}} &\textbf{IC} & &6 &8 & &100 &98 \\
&\textbf{ML} & &8 &6 & &68 &50 \\
\multirow{2}{*}{\textbf{20}} &\textbf{IC} & &0 &0 & &0 &0 \\
&\textbf{ML} & &0 &0 & &100 &98 \\
\multirow{2}{*}{\textbf{50}} &\textbf{IC} & &0 &0 & &80 &72 \\
&\textbf{ML} & &0 &0 & &0 &0 \\
\bottomrule
\end{tabular}}
\end{table*}

\section{Generating Stable Solutions for Preferences with Ties} \label{app:ties}
As demonstrated in \Cref{sec:generating}, reasoning-enabled models achieve significantly higher accuracy on Easy instances when compared to non-reasoning baseline models, but suffer dramatic drops in performance with Hard instances. A natural extension of this is to examine how introducing ties to preferences impacts an LLM's ability to generate stable solutions.

\paragraph{New notions of stability.} When ties are introduced to preference profiles, additional (stronger) notions of stability exist. Namely, in addition to the standard notion of \textit{weak stability} (where a matching does not admit any weak blocking pairs: agents who strictly prefer each other to their current partners), we have the added notions of \textit{strong} and \textit{super stability}, where a matching does not admit any strong or super blocking pairs, respectively. A strong blocking pair is a pair of agents in which one agent strictly prefers the other agent over their current partner, and the other agent remains indifferent or prefers the first agent and their current partner. A super blocking pair is a pair of agents in which either agent is either indifferent the other agent and their current partner, or prefers the other agent to their current partner. \textit{Super stability}, the strongest notion of stability, implies \textit{strong stability}, which in turn implies \textit{weak stability}. Additionally, it is important to note that \textit{strongly} and \textit{super stable} solutions do not exist for all preference profiles.

Irving \cite{IRVING1994261} provides three algorithms (one for each) to compute \textit{weakly}, \textit{strongly}, and \textit{super stable} matchings for preference profiles with ties. The algorithm for \textit{weak} stability takes $\mathcal{O}(n^2)$ steps (the algorithm is equivalent to running standard the DA algorithm with arbitrary tie-breaking), the algorithm for \textit{strong} stability takes $\mathcal{O}(n^4)$ steps, and the algorithm for \textit{super} stability takes $\mathcal{O}(n^2)$ steps (these two algorithms require more demanding steps with complex operations).

\begin{figure}[t]
    \centering
    \includegraphics[width=0.6\linewidth]{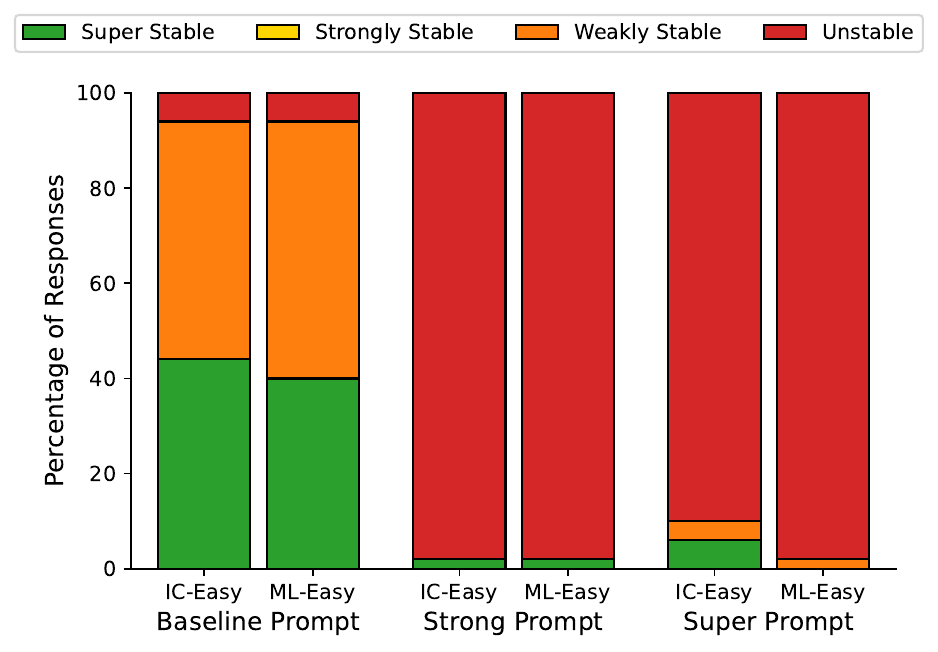}
    \caption{The generated responses by \GemPro{} for Master-list (ML) and Impartial Culture (IC) Easy preferences with different prompts. All generated matchings were valid and complete. Again, \textit{unstable} indicates one-to-one matchings with \textit{weak} blocking pairs present. 
    }
    \label{fig:strong_super}
\end{figure}

\paragraph{Generating matchings for each stability notion.}
To evaluate how LLMs handle preferences with ties, we perform the same experiment as
in \Cref{sec:generating} with Easy preference profiles and \GemPro, except we introduce  multiple ties of random sizes, at randomly selected starting positions, in each preference list in the profile.
Additionally, we modify the prompt listed in \Cref{app:gen-stable-sol-vanilla} to get a total of three different prompts. The first prompt (referred to as the \textit{Baseline Prompt}) is identical to the prompt in \Cref{app:gen-stable-sol-vanilla}. The other two prompts (referred to as the \textit{Strong Prompt} and \textit{Super Prompt}) replace the instruction to ``...find the proposer-optimal stable matching...'' with ``...find the proposer-optimal STRONG stable matching...'' and ``...find the proposer-optimal SUPER stable matching...'', respectively.
For each prompt, we run the experiment and count the proportion of generated matchings that are \textit{super}, \textit{strong}, and \textit{weakly stable}, as well as unstable matchings. 

\Cref{fig:strong_super} demonstrates the performance of \GemPro{} in generating stable outcomes with respect to each of the three new notions of stability for preference profiles with ties. \GemPro{} once again demonstrates a high accuracy ($96\%$)
in generating stable solutions (considering \textit{weak}, \textit{strong}, and \textit{super stability}) with the Baseline Prompt, with $50\%$ and $54\%$ of stable solutions being \textit{weakly stable} with the IC-Easy and ML-Easy preference profiles, respectively. However, only around $40\%$ of generated solutions were super (and strongly) stable for both preference cultures. 
\GemPro{}'s reasoning traces for each instance indicate that the model essentially attempts to execute the DA algorithm when generating stable solutions even when ties are present  (rather than utilizing the more complex algorithms for \textit{strong} and \textit{super stability}, even when explicitly prompted). Additionally, any generated \textit{super} stable solutions are a result of an instance's \textit{super} stable solution intersecting with its \textit{weakly} stable solution. Interestingly, specifying the notion of stability appears to significantly hurt the model's ability to generate any stable solutions. While one potential explanation is that specifying the desired stability notion in the prompt introduces unnecessary noise, this is an interesting avenue for future study. 

\section{Resolving Instability: Additional Material} \label{app:resolving}

\subsection{Generating Unstable Matchings}
Here we describe the procedures we use to generate the two types of unstable matchings we consider.

\paragraph{One-BP.} We generate a matching that contains a single blocking pair, by swapping the partners of two randomly selected proposer agents in the Optimal matching. Since such a swap may lead to more than one blocking-pair (or no blocking pairs), we perform this process (for every instance) until we obtain a matching with exactly one blocking-pair. This procedure is formally described in \Cref{alg:generate-unstable}.

\begin{algorithm}[ht]
  \caption{\textsc{GenerateOneBPMatching}}
  \label{alg:generate-unstable}
  \begin{algorithmic}[1]
    \Require
      \Statex $\,\Pi = (\succ_m, \succ_w)$ \Comment{Preference profile for all men $m\in M$ and women $w\in W$}
      \Statex $\mu^\ast$ \Comment{Men-optimal stable matching returned by Deferred Acceptance}
    \Ensure
      \Statex A matching $\mu$ containing \emph{exactly one} blocking pair
    \Function{GenerateOneBPMatching}{$\Pi, \mu^\ast$}
      \Repeat                                           \Comment{Keep trying until the condition is met}
        \State $\mu \gets \text{copy}(\mu^\ast)$        \Comment{Start from the stable matching}
        \State $(m_a, m_b) \gets \text{arbitrary pair } m_a, m_b \in M \text{ s.t. } m_a \neq m_b$
        \State $w_a \gets \mu(m_a)$
        \State $w_b \gets \mu(m_b)$
        \Comment{Swap partners of the two men}
        \State $\mu(m_a) \gets w_b,\;\; \mu(w_b) \gets m_a$
        \State $\mu(m_b) \gets w_a,\;\; \mu(w_a) \gets m_b$
      \Until{$\lvert $\textsc{BlockingPairs}$(\mu,\Pi)\rvert = 1$}          \Comment{Stop when exactly one blocking pair exists}
      \State \Return $\mu$
    \EndFunction
    \bigskip
    \Function{BlockingPairs}{$\mu,\Pi$}
      \State $B \gets \emptyset$
      \ForAll{$m \in M$}
        \ForAll{$w \in W$}
          \If{$w \succ_m \mu(m)$ \textbf{and} $m \succ_w \mu(w)$}
            \State $B \gets B \cup \{(m,w)\}$
          \EndIf
        \EndFor
      \EndFor
      \State \Return $B$
    \EndFunction
  \end{algorithmic}
\end{algorithm}

\paragraph{Random.} A Random matching is simply generated by generating a random permutation of agents on one side and assigning agents such a list to the agents on the other side, one-by-one.

\subsection{Further Results}

\begin{table*}[t]
    \centering
    \caption[]{
    Instability Rate (averaged across instances) in the (valid) matchings returned by LLMs when asked to resolve a given unstable matching of types One-BP and Random. The column ``Baseline'' contains the (average) Instability Rate for the provided matching of the indicated type. 
    Numbers in bold indicate that the Instability Rate of the corrected solution is significantly worse than the provided matching. A $*$ on the number in the One-BP column indicates a that Instability Rate is significantly lower than the case when a Random matching is provided (at $p < 0.05$).\footnotemark
    }
    \resizebox{\textwidth}{!}{
    \begin{tabular}{lc*{23}{c}}
        \toprule
        & \multicolumn{5}{c}{\textbf{Basic LLMs}} & & \multicolumn{5}{c}{\textbf{Reasoning LLMs}} & & \multicolumn{8}{c}{\textbf{Advanced Reasoning LLMs}} \\\cmidrule{2-6}\cmidrule{8-12}\cmidrule{14-21}
        & \multicolumn{2}{c}{\textbf{\GemFlash}} &
        & \multicolumn{2}{c}{\textbf{\Llama}} & 
        & \multicolumn{2}{c}{\textbf{\QwQ}} &
        & \multicolumn{2}{c}{\textbf{\DSD}} &
        & \multicolumn{2}{c}{\textbf{\othree}} &
        & \multicolumn{2}{c}{\textbf{\Deepseek}} &
        & \multicolumn{2}{c}{\textbf{\GemPro}} & 
        & \multicolumn{2}{c}{\textbf{Baseline}} 
        \\\cmidrule{2-3}\cmidrule{5-6}\cmidrule{8-9}\cmidrule{11-12}\cmidrule{14-15}\cmidrule{17-18}\cmidrule{20-21}\cmidrule{23-24}
        \textbf{Difficulty}
        & \textbf{One-BP} & \textbf{Random} &
        & \textbf{One-BP} & \textbf{Random} &
        & \textbf{One-BP} & \textbf{Random} &
        & \textbf{One-BP} & \textbf{Random} &
        & \textbf{One-BP} & \textbf{Random} &
        & \textbf{One-BP} & \textbf{Random} &
        & \textbf{One-BP} & \textbf{Random} &
        & \textbf{One-BP} & \textbf{Random} 
        \\
        \midrule
        \textbf{Easy} & $\boldsymbol{45.5}$ & $46.3$ & & $\boldsymbol{34.25^*}$ & $42.05$ & & $4.30^*$ & $10.80$ & & $8.74$ & $9.05$ & & $0.25$ & $0.00$ & & $0.20$ & $0.20$ & & $0.30$ & $0.80$ 
        & & $10.00$ & $77.32$ 
        \\\midrule
        \textbf{Medium} & $\boldsymbol{55.09}$ & $58.81$ & & $\boldsymbol{57.48^*}$ & $64.12$ & & $\boldsymbol{14.50^*}$ & $49.20$ & & $\boldsymbol{23.39^*}$ & $43.84$ & & $3.92$ & $3.25$ & & $\boldsymbol{10.65}$ & $12.58$ & & $2.17$ & $3.5$
        & & $5.00$ & $87.95$ 
        \\\midrule
        \textbf{Hard} & $\boldsymbol{49.99^*}$ & $74.66$ & & $\boldsymbol{61.42^*}$ & $86.79$ & & $\boldsymbol{17.44^*}$ & $86.91$ & & $\boldsymbol{35.38^*}$ & $84.76$ & & $\boldsymbol{43.77^*}$ & $59.98$ & & $\boldsymbol{14.8^*}$ & $47.81$ & & $\boldsymbol{20.99^*}$ & $25.08$ 
        & & $2.00$ & $94.52$
        \\
        \bottomrule
        \end{tabular}
        }
    \label{tab:resolve_bp}
\end{table*}

\footnotetext{All statistical comparisons in this table are made using Welch's t-test \citep{WelchTTest}.}

\begin{figure}[t]
    \centering
    \includegraphics[width=1\linewidth]{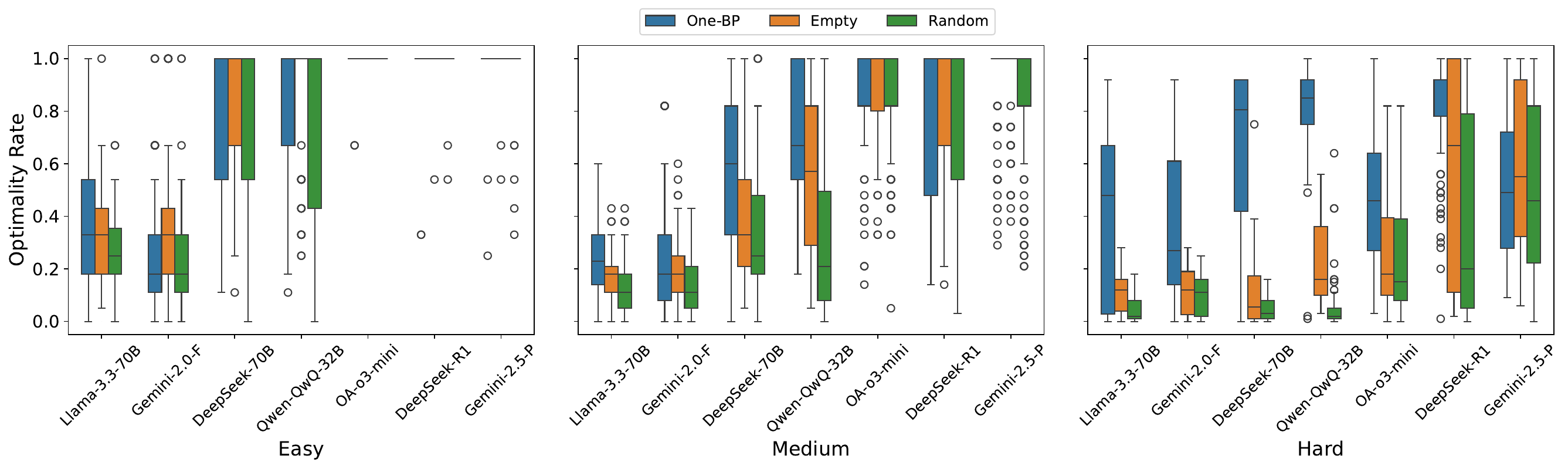}
    \caption{Optimality Rate when LLMs are asked to resolve a matching with a single blocking pair introduced into the Optimal solution (One-BP), or a randomly generated matching (Random). These are is compared to the case when they are asked to generate a stable matching from scratch (Empty).
    }
    \label{fig:resolve_intersections}
\end{figure}

\paragraph{Measuring Instability after Repair.} The extent to which a given matching is incorrect does influence quality of the solutions returned by LLMs after being asked to correct it. \Cref{tab:resolve_bp} shows that the number of blocking pairs is often significantly lower when the unstable matching (that LLMs are asked to correct) has a single blocking pair, as compared to the case with random matchings.\footnote{In fact, asking LLMs to resolve a random matching leads to a significantly higher number of blocking pairs in the returned solution, as compared to the case when they are asked to generate solutions from scratch.} Similarly, as shown in \Cref{fig:resolve_intersections}, matchings returned after resolving an almost stable matching have a greater overlap with the Optimal solution (especially at Medium and Hard difficulty levels).

\section{Additional Material about Detecting Instability}\label{app:detecting}

\subsection{Comparing Impartial Culture and Master-List Instances}
Generally, the classification of a preference profile as an impartial culture or Master-list instance has a relatively small impact on the ability of an LLM to detect instability in the instance. However, we observe some differences between the ability of certain LLMs to detect stable/unstable matchings with ML instances compared to IC instances. Models such as \DSQwenMed, \othree, and \DSD{} are able correctly detect stable solutions significantly more frequently with ML instances than with IC instances. A potential explanation for this is that Master-list preferences contain fewer \textit{unique} preference lists, decreasing the chances that the model hallucinates blocking pairs. On the other hand, models such as \GemFlash{} and \Llama{} correctly identify unstable solutions significantly more often with IC preferences as compared to the case with ML preferences. The intuition for this observation is the opposite: with impartial culture preferences, there is a higher probability of having blocking pairs, therefore models that tend to predict that solutions are unstable will perform better with impartial culture instances.

\section{Fine-tuning Details}\label{app:ft_details}

\paragraph{Models.} We fine-tune four reasoning models:
\begin{itemize}
    \item \DSLlamaSmall{} (\texttt{deepseek-ai/DeepSeek-R1-Distill-Llama-8B}),
    \item \DSQwenMed{} (\texttt{deepseek-ai/DeepSeek-R1-Distill-Qwen-14B}),
    \item \QwQ{} (\texttt{Qwen/QwQ-32B}), and 
    \item \Qwen{} (\texttt{Qwen/Qwen3-32B})
\end{itemize}
using the Unsloth\footnote{https://unsloth.ai/} framework with parameter-efficient tuning (LoRA).\footnote{Since results for \QwQ{} and \Qwen{} are similar, we show only those for the former in \Cref{tab:ft_enhancement}.}

\paragraph{Dataset.} The dataset for the Generation task contains $N=10000$ samples for which the reasoning trace is generated using a Python implementation of the DA algorithm. For the Preference Reasoning task, the dataset consists of $N=9000$ samples ($3000$ for each question level), where the reasoning trace involves explicitly identifying the positions of agents in the concerned preferences. In both datasets, we include an equal number of IC and ML instances, with sizes ranging from $n=5$ to $n=50$. Detailed examples of training examples for both tasks (Generation and Preference Reasoning) are provided in \Cref{app:ft_seqs}.

\paragraph{Model Setup.} We used the \texttt{FastLanguageModel.from\_pretrained} interface from Unsloth to load the base model with a maximum sequence length of 10,000 tokens. The model was loaded in full precision (no quantization) and fine-tuned using Low-Rank Adaptation (LoRA) with the following settings:
\begin{itemize}
    \item Rank ($r$): 32
    \item Target Modules: \texttt{q\_proj}, \texttt{k\_proj}, \texttt{v\_proj}, \texttt{o\_proj}, \texttt{gate\_proj}, \texttt{up\_proj}, \texttt{down\_proj}
    \item LoRA $\alpha$: 32
    \item LoRA Dropout: 0
    \item Bias: \texttt{none}
    \item Gradient Checkpointing: Enabled via \texttt{use\_gradient\_checkpointing="unsloth"}
\end{itemize}

\paragraph{Training Configuration.} Fine-tuning was conducted using the \texttt{SFTTrainer} from the TRL library with the following training arguments:
\begin{itemize}
    \item Epochs: 1
    \item Batch size per device: 2 (1, for \QwQ{})
    \item Gradient accumulation steps: 4 (2, for \QwQ{})
    \item Learning rate: $2 \times 10^{-4}$ with a linear scheduler and 5 warmup steps
    \item Optimizer: \texttt{AdamW-8bit}
    \item Weight decay: 0.01
    \item Precision: Mixed precision (FP16 or BF16, based on hardware support)
    \item Seed: 3407
\end{itemize}

\paragraph{Hardware.} Each model was fine-tuned using a single NVIDIA H100 GPU (80GB RAM) with CUDA support; model and inputs were explicitly transferred to GPU for inference and training.

\paragraph{Model Saving and Sharing.} The resulting models were uploaded to the Hugging Face Hub and will be released upon acceptance.

\paragraph{Inference Setup.} After fine-tuning, the model was evaluated using in-context inference. Inputs were formatted similarly to training prompts, and the model's output was parsed to extract the JSON-formatted matching solution.

\paragraph{Ablation tests.} \Cref{tab:ft_ablations} illustrates how performance is sensitive to variations in data-related parameters such as the types of instances included in the data, range of instance sizes, and number of training examples, as well as training-related parameters such as the LoRA rank and the base model used. The primary observation from these tests is that easier training instances are more crucial that harder ones. For example, performance is significantly better when the training data consists of only Easy (and smaller) instances, i.e. with $n = 10$, as compared to the case with only Hard instances ($n = 50$). Similarly, performance is much better when the data consists of only ML instances, as opposed to the case where it consists of only IC instances, which are more difficult than the former.

\begin{table*}\centering
\caption{Performance (percentage of stable solutions generated) after fine-tuning, for \DSLlamaSmall{} and \Qwen{}, with different configurations of data- and training-related hyper-parameters. Configuration 1 is the default configuration. The variation in each other configuration (as compared to config. 1) is in bold.}
\label{tab:ft_ablations}
\resizebox{\textwidth}{!}{ 
\begin{tabular}{ccccccccccccccc}\toprule
& & & & & &\multicolumn{3}{c}{\textbf{ML}} & &\multicolumn{3}{c}{\textbf{IC}} & \\\cmidrule{7-9}\cmidrule{11-13}
\textbf{Config.} &\textbf{\makecell{Instance\\types}} &\textbf{\makecell{Instance sizes\\(range)}} &\textbf{\makecell{Training\\set size}} &\textbf{\makecell{LoRA\\rank}} &\textbf{Base model} &\textbf{Easy} &\textbf{Medium} &\textbf{Hard} & &\textbf{Easy} &\textbf{Medium} &\textbf{Hard} &\textbf{Total} \\\midrule
\multirow{2}{*}{\textbf{1}} &\multirow{2}{*}{ML, IC} &\multirow{2}{*}{[5,50]} &\multirow{2}{*}{10000} &\multirow{2}{*}{32} &\textbf{DeepSeek-8B} &94 &60 &0 & &34 &2 &0 &31.67 \\\cmidrule{6-14}
& & & & &\textbf{Qwen3-32B} &100 &98 &4 & &98 &84 &0 &64 \\\midrule
\multirow{2}{*}{\textbf{2}} &\multirow{2}{*}{\textbf{ML}} &\multirow{2}{*}{[5,50]} &\multirow{2}{*}{10000} &\multirow{2}{*}{32} &\textbf{DeepSeek-8B} &100 &94 &0 & &2 &0 &0 &32.67 \\\cmidrule{6-14}
& & & & &\textbf{Qwen3-32B} &100 &100 &0 & &6 &0 &0 &34.33 \\\midrule
\multirow{2}{*}{\textbf{3}} &\multirow{2}{*}{\textbf{IC}} &\multirow{2}{*}{[5,50]} &\multirow{2}{*}{10000} &\multirow{2}{*}{32} &\textbf{DeepSeek-8B} &0 &0 &0 & &0 &0 &0 &0 \\\cmidrule{6-14}
& & & & &\textbf{Qwen3-32B} &0 &0 &0 & &0 &2 &0 &0.33 \\\midrule
\multirow{2}{*}{\textbf{4}} &\multirow{2}{*}{ML, IC} &\multirow{2}{*}{[5,50]} &\multirow{2}{*}{10000} &\multirow{2}{*}{\textbf{64}} &\textbf{DeepSeek-8B} &96 &82 &0 & &2 &0 &0 &30 \\\cmidrule{6-14}
& & & & &\textbf{Qwen3-32B} &100 &100 &0 & &100 &100 &0 &66.67 \\\midrule
\multirow{2}{*}{\textbf{5}} &\multirow{2}{*}{ML, IC} &\multirow{2}{*}{[5,50]} &\multirow{2}{*}{\textbf{5000}} &\multirow{2}{*}{32} &\textbf{DeepSeek-8B} &74 &46 &0 & &0 &0 &0 &20 \\\cmidrule{6-14}
& & & & &\textbf{Qwen3-32B} & 100 & 94 & 0 & & 98 & 92 & 0 &64 \\\midrule
\multirow{2}{*}{\textbf{6}} &\multirow{2}{*}{ML, IC} &\multirow{2}{*}{\textbf{[5,10]}} &\multirow{2}{*}{\textbf{5000}} &\multirow{2}{*}{32} &\textbf{DeepSeek-8B} & 100 & 2 & 0 & & 100 & 8 & 0 &35 \\\cmidrule{6-14}
& & & & &\textbf{Qwen3-32B} & 100 & 52 & 0 & & 100 & 26 & 0 & 46.33 \\\midrule
\multirow{2}{*}{\textbf{7}} &\multirow{2}{*}{ML, IC} &\multirow{2}{*}{\textbf{[50,50]}} &\multirow{2}{*}{\textbf{5000}} &\multirow{2}{*}{32} &\textbf{DeepSeek-8B} &0 &0 &0 & &0 &0 &0 &0 \\\cmidrule{6-14}
& & & & &\textbf{Qwen3-32B} &12 &0 &0 & &2 &0 &0 &2.33 \\
\bottomrule
\end{tabular}}
\end{table*}

\section{Inference Details}\label{app:inference}

\paragraph{Fine-tuned Models.} The models we fine-tune have been pushed to HuggingFace Hub and will be released upon acceptance.

\paragraph{Inference Configuration.} For the task of generating stable solutions, inference was performed on each of the open-source models such as \DSLlamaSmall{}, \DSQwenMed{}, \QwQ{}, \DSD{}, and \Llama, with the following sampling parameters:
\begin{itemize}
    \item Temperature: 0.5 
    \item Maximum tokens: 30,000 
\end{itemize}
Default values were utilized for all other sampling parameters. 
We used online APIs for the following models:
\begin{itemize}
    \item \GemFlash{} ('gemini-2.0-flash')
    \item \othree{} (`o3-mini')
    \item \Deepseek{} (`deepseek-reasoner')
    \item \GemPro{} (`gemini-2.5-pro-preview-03-25')
\end{itemize}

\paragraph{Hardware.} All inference experiments with open-source models were run on NVIDIA H100 GPUs (80GB RAM) with CUDA support; model and inputs were explicitly transferred to GPU for inference and training. We used a single GPU for inference involving \DSLlamaSmall{} and \DSQwenMed{}, two GPUs for inference involving \QwQ{}, and four GPUs for inference involving \Llama{} and \DSD{}.

\newpage
\section{Prompts}\label{app:prompts}

\subsection{Example prompt for Generating Stable Solutions}\label{app:gen-stable-sol}

\paragraph{Vanilla prompt.} \label{app:gen-stable-sol-vanilla} The prompt used for generating stable solutions with LLMs follows standard prompting procedures, by first outlining the task, providing appropriate context, specifying constraints, and detailing the desired output format (a JSON object). We intentionally provide preferences in a structured, tabular format. It enables us to isolate and rigorously evaluate the LLMs' reasoning, alignment, and solution quality relative to normative axioms (e.g., stability). Natural language formulations introduce significant noise in both input and output, making it difficult to attribute performance failures to reasoning versus parsing. By grounding our analysis in tabular settings first, we can obtain clean and interpretable measurements, forming a benchmark for future extensions that incorporate naturalistic input.

Notice that despite the deferred-acceptance algorithm never being mentioned in the prompt, all models mentioned the deferred-acceptance algorithm in their responses. As mentioned in \Cref{app:CoT}, we use the traditional setting of stable-marriage (where men propose to women) considered by \citet{gale1962college} while describing the problem in the prompt.

\begin{tcolorbox}[fontupper=\sffamily]
You are an intelligent assistant who is an expert in algorithms. Consider the following instance of the two-sided matching problem, where 10 men are to be matched with 10 women. 
Here are the preference lists for all individuals:

$<$preferences$>$

\{

M: \{

M1: [W10,W1,W3,W6,W2,W4,W9,W8,W7,W5],

M2: [W8,W3,W10,W6,W2,W5,W4,W7,W1,W9],

...

M10: [W2,W5,W1,W3,W7,W6,W10,W4,W9,W8],

\},

W: \{

W1: [M2,M8,M9,M10,M5,M7,M1,M4,M6,M3],

W2: [M2,M7,M3,M1,M8,M9,M6,M10,M5,M4],

...

W10: [M6,M4,M7,M5,M8,M9,M10,M2,M3,M1],

\}\}

$<$/preferences$>$

Your task is to find the proposer-optimal stable matching. You can use XML tags like $<$scratchpad$>$ to explain your thought process while computing the solution.

Once you have found a stable matching, please return your matching in the JSON format given below:

$<$answer$>$

\{

        ``M1'': ``$<$woman matched with M1$>$'',
        
        ``M2'': ``$<$woman matched with M2$>$'',

        ...
        
        ``M10'': ``$<$woman matched with M10$>$''

\}

$<$/answer$>$

Make sure that each man/woman is matched with exactly ONE partner. It is mandatory that you provide a matching as a JSON object enclosed in $<$answer$>$$<$/answer$>$ tags as described above.
\end{tcolorbox}

\newpage

\paragraph{Providing Algorithmic Description.} The following is the prompt is a modification of the vanilla prompt where the steps of the DA algorithm have been described to assist the model with implementing the same.

\begin{tcolorbox}[fontupper=\sffamily]
You are an intelligent assistant who is an expert in algorithms. 

...

$<$/preferences$>$

Your task is to find the proposer-optimal stable matching.  For this, you can use the Deferred Acceptance algorithm. The steps of this algorithm are described below:

1. Initialize all men and women as unmatched.

2. Create a list to keep track of each man's next proposal (initially set to 0 for all men).

3. While there are unmatched men:

   a. Select an unmatched man (M).

   b. Find the next woman (W) on M's preference list that he hasn't proposed to yet.

   c. If W is unmatched, match M and W.

   d. If W is matched but prefers M to her current partner:

      - Unmatch W from her current partner.

      - Match M and W.

      - Set the unmatched man as W's previous partner.

   e. If W rejects M, move to the next woman on M's preference list.

4. Repeat step 3 until all men are matched.

You can use XML tags like $<$scratchpad$>$ to explain your thought process ...

...

It is mandatory that you provide a matching as a JSON object enclosed in $<$answer$>$$<$/answer$>$ tags as described above.
\end{tcolorbox}

\paragraph{Modified Problem Setting.} The following is a modification to the vanilla prompt, where the setting of \textit{task-allocation} (assigning tasks to workers) is considered instead of the \textit{stable-marriage} setting. We replace \textit{men} with \textit{workers} and \textit{women} with \textit{tasks}.

\begin{tcolorbox}[fontupper=\sffamily]
You are an intelligent assistant who is an expert in algorithms. Consider the following instance of the two-sided matching problem, where 5 workers are to be assigned with 5 tasks, and each worker is assigned exactly one task. 

Here are the preference lists for all workers (W) over tasks (T) and the preferences of tasks over workers:

$<$preferences$>$

\{

W: \{

W1: [T5, T3, T4, T2, T1]

...

\}

T: \{

T1: [W3, W5, W4, W1, W2]

...

\}\}

$<$/preferences$>$

Your task is to find a stable matching of workers and tasks. You can use XML tags like $<$scratchpad$>$ to explain your thought process while computing the solution.

Once you have found a stable matching, please return your matching in the JSON format given below:

$<$answer$>$

\{

        "W1": "$<$task assigned to W1$>$",

        ...

        "W5": "$<$task assigned to W5$>$"

\}

$<$/answer$>$

Make sure that each worker is assigned exactly ONE task. It is mandatory that you provide a matching as a JSON object enclosed in $<$answer$>$$<$/answer$>$ tags as described above.
\end{tcolorbox}

\subsection{Example Prompts for Prompt Engineering}

\subsubsection{CoT-Vanilla}\label{app:cot-vanilla}

Chain-of-Thought methods were applied to the prompt in \Cref{app:gen-stable-sol} by additionally including an example trace of steps performed when running the deferred-acceptance algorithm on a randomly generated instance. The algorithm trace includes all proposals, all respective acceptances/rejections, and the resultant stable solution. The entire Chain-of-Thought example is enclosed within $<$example$>$ XML tags.

\begin{tcolorbox}[fontupper=\sffamily]
You are an intelligent assistant who is an expert in algorithms. Your task is to find the proposer-optimal stable matching, for the two-sided matching problem. Here is an example to demonstrate how you should proceed:

$<$example$>$

$<$preferences$>$

\{

M: \{

M1: [W5,W1,W2,W4,W3],

M2: [W1,W2,W5,W4,W3],

M3: [W4,W2,W3,W1,W5],

M4: [W5,W1,W2,W4,W3],

M5: [W3,W5,W4,W2,W1],

\},

W: \{

W1: [M2,M3,M5,M4,M1],

W2: [M5,M2,M4,M3,M1],

W3: [M2,M1,M3,M5,M4],

W4: [M1,M4,M5,M3,M2],

W5: [M4,M3,M5,M2,M1],

\}\}

$<$/preferences$>$

M1 is free. M1 proposes to W5

Since W5 is free, W5 accepts the proposal. Now M1 and W5 are matched.

M2 is free. M2 proposes to W1

Since W1 is free, W1 accepts the proposal. Now M2 and W1 are matched.

M3 is free. M3 proposes to W4

Since W4 is free, W4 accepts the proposal. Now M3 and W4 are matched.

M4 is free. M4 proposes to W5

Since W5 prefers M4 to their current partner M1, W5 accepts the proposal. Now M4 and W5 are matched, and M1 is free.

M1 is free. M1 proposes to W1

Since W1 prefers their current partner M2 to M1, W1 rejects the proposal. M2 and W1 are still matched, and M1 is still free.

M1 is free. M1 proposes to W2

Since W2 is free, W2 accepts the proposal. Now M1 and W2 are matched.

M5 is free. M5 proposes to W3

Since W3 is free, W3 accepts the proposal. Now M5 and W3 are matched.

$<$answer$>$

\{

        "M1": "W2",

        "M2": "W1",

        "M3": "W4",

        "M4": "W5",

        "M5": "W3"

\}

$<$/answer$>$

$<$/example$>$

Consider the following instance of the two-sided matching problem, where 10 men are to be matched with 10 women $\ldots$

\end{tcolorbox}

\newpage

\subsubsection{CoT-Shortlist}
The main distinction between the prompt described here and the one in \Cref{app:cot-vanilla} lies in how the algorithm's execution is detailed. In the CoT-Shortlist prompt, the provided algorithm trace includes an additional step: agents remove each other from their respective shortlists if they become matched with a partner they find more desirable than the other agents on their list. All other aspects of the prompt are identical to the CoT-Vanilla prompt.

\begin{tcolorbox}[fontupper=\sffamily]
You are an intelligent assistant who is an expert in algorithms. Your task is to find the proposer-optimal stable matching, for the two-sided matching problem. Here is an example to demonstrate how you should proceed:

$<$example$>$

$<$preferences$>$

\{

M: \{

M1: [W4,W3,W5,W2,W1],

M2: [W5,W4,W3,W1,W2],

M3: [W5,W4,W1,W2,W3],

M4: [W5,W4,W2,W1,W3],

M5: [W2,W4,W5,W3,W1],

\},

W: \{

W1: [M5,M2,M3,M4,M1],

W2: [M3,M4,M5,M1,M2],

W3: [M4,M1,M2,M5,M3],

W4: [M5,M1,M4,M3,M2],

W5: [M1,M4,M5,M3,M2],

\}\}

$<$/preferences$>$

M1 is free. M1 proposes to W4. W4 accepts the proposal. Now M1 and W4 are matched.

W1 deletes M4, M3, M2 from her list. M4, M3, M2 delete W4 from their list.

M2 is free. M2 proposes to W5. W5 accepts the proposal. Now M2 and W5 are matched.

M3 is free. M3 proposes to W5. W5 accepts the proposal. Now M3 and W5 are matched.

W5 prefers M3, so W5 breaks her engagement with M2.

W3 deletes M2 from her list. M2 delete W5 from their list.

M4 is free. M4 proposes to W5. W5 accepts the proposal. Now M4 and W5 are matched.

W5 prefers M4, so W5 breaks her engagement with M3.

W4 deletes M5, M3 from her list. M5, M3 delete W5 from their list.

M5 is free. M5 proposes to W2. W2 accepts the proposal. Now M5 and W2 are matched.

W5 deletes M1, M2 from her list. M1, M2 delete W2 from their list.

$<$answer$>$

\{

        "M1": "W4",

        "M2": "W3",

        "M3": "W1",

        "M4": "W5",

        "M5": "W2"

\}

$<$/answer$>$

$<$/example$>$

Consider the following instance of the two-sided matching problem, where 10 men are to be matched with 10 women $\ldots$

\end{tcolorbox}
\newpage

\subsubsection{Few-shot Examples}

Few-shot prompting was applied to the prompt in \Cref{app:gen-stable-sol} by additionally including a series of randomly generated preference/stable solution pairs. As with other few-shot prompting strategies, the model is then asked to generate a stable solution (as shown in \Cref{app:gen-stable-sol}). As with the CoT methods, each sample preference/stable solution pairs is enclosed in $<$example$>$ XML tags.

\begin{tcolorbox}[fontupper=\sffamily]
You are an intelligent assistant who is an expert in algorithms. Your task is to find the proposer-optimal stable matching, for the two-sided matching problem. Here is an example to demonstrate how you should proceed:

$<$example$>$

$<$preferences$>$

\{

M: \{

M1: [W5,W3,W4,W2,W1],

M2: [W3,W4,W1,W2,W5],

M3: [W5,W1,W4,W2,W3],

M4: [W3,W2,W5,W1,W4],

M5: [W3,W4,W2,W1,W5],

\},

W: \{

W1: [M1,M4,M3,M5,M2],

W2: [M2,M4,M5,M1,M3],

W3: [M1,M2,M4,M5,M3],

W4: [M3,M5,M1,M4,M2],

W5: [M5,M3,M4,M2,M1],

\}\}

$<$/preferences$>$

$<$answer$>$

\{

        "M1": "W3",

        "M2": "W1",

        "M3": "W5",

        "M4": "W2",

        "M5": "W4"

\}

$<$/answer$>$

$<$/example$>$

$<$example$>$

...

$<$/example$>$

$<$example$>$

...

$<$/example$>$

Consider the following instance of the two-sided matching problem, where 10 men are to be matched with 10 women $\ldots$

\end{tcolorbox}

\subsection{Example Prompt for Evaluating Stability}

The following prompt requires LLMs to determine if a given solution to a provided preference profile is stable. Unlike the prompt in \Cref{app:gen-stable-sol}, the only element that the LLM must include in their response is a binary response (yes/no).
\begin{tcolorbox}[fontupper=\sffamily]
    Consider the following instance of the two-sided matching problem, where 5 men are to be matched with 5 women. 

Here are the preference lists for all individuals:

$<$preferences$>$

\{

M: \{

M1: [W5,W3,W4,W2,W1],

...

\},

W: \{

W1: [M3,M5,M4,M1,M2],

...

\}\}

$<$/preferences$>$

Your task is to determine whether the following matching is stable or not.

$<$matching$>$

[[M1, W4],[M2, W5],[M3, W3],[M4, W1],[M5, W2],]

$<$/matching$>$

Please return 'Yes' if you think the provided matching is stable and 'No' if you think it is unstable, and enclose your answer in $<$answer$>$$<$/answer$>$ tags.
\end{tcolorbox}

\subsection{Example Prompts for Preference Comprehension}

In each of the following preference comprehension prompts, models are asked to provide the name of an agent (in level-1) or to provide a binary answer (yes/no for levels 2 and 3) in response to a provided question. In addition to changing the preference profiles for each instance of a preference comprehension task, the agents and positions mentioned in the question are also changed with each instance. For details about each level of preference comprehension, view \Cref{sec:reasoningPref}.

\subsubsection{Level-1}

\begin{tcolorbox}[fontupper=\sffamily]
    Your goal is to correctly interpret the given preference lists and answer a specific question about agent preferences.

First, here are the preference lists for all individuals:

$<$preferences$>$

\{

M: \{

M1: [W5,W3,W4,W2,W1],

...

\},

W: \{

W1: [M3,M5,M4,M1,M2],

W2: [M1,M3,M4,M5,M2],

...

\}\}

$<$/preferences$>$

Now, you will be asked a specific question about agent preferences:

$<$question$>$

Who is agent W2's, 1-most preferred agent?

$<$/question$>$

Once you have determined the answer, provide your output in the following format:

1. The solution as a single agent name. For example, "W1"

Present your final answer within $<$answer$>$ tags.

IMPORTANT: ONLY RETURN THE NAME OF THE SINGLE AGENT THAT IS THE ANSWER TO THE QUESTION. Do not include any explanations or additional information in your final answer.
\end{tcolorbox}

\subsubsection{Level-2}

\begin{tcolorbox}[fontupper=\sffamily]
    You are an AI assistant tasked with analyzing preference profiles in a two-sided matching problem with one-to-one solutions. Your goal is to correctly interpret the given preference lists and answer a specific question about agent preferences.

First, here are the preference lists for all individuals:

$<$preferences$>$

\{

M: \{

M1: [W5,W3,W4,W2,W1],

...

\},

W: \{

W1: [M3,M5,M4,M1,M2],

...

\}\}

$<$/preferences$>$

Now, you will be asked a specific question about agent preferences:

$<$question$>$

Would agent W1, prefer M3 or M2 over M4?

$<$/question$>$

Once you have determined the answer, provide your output in the following format:

1. The solution as a YES or a NO. For example, "NO"

Present your final answer within $<$answer$>$ tags.

IMPORTANT: ONLY RETURN YES OR NO THAT IS THE ANSWER TO THE QUESTION. Do not include any explanations or additional information in your final answer.
\end{tcolorbox}

\subsubsection{Level-3}

\begin{tcolorbox}[fontupper=\sffamily]
    You are an AI assistant tasked with analyzing preference profiles in a two-sided matching problem with one-to-one solutions. Your goal is to correctly interpret the given preference lists and answer a specific question about agent preferences.

First, here are the preference lists for all individuals:

$<$preferences$>$

\{

M: \{

M1: [W5,W3,W4,W2,W1],

...

\},

W: \{

W1: [M3,M5,M4,M1,M2],

...

\}\}

$<$/preferences$>$

Now, you will be asked a specific question about agent preferences:

$<$question$>$

If agent W1 is currently engaged to M4, would she accept proposals from M3 or M2?

$<$/question$>$

Once you have determined the answer, provide your output in the following format:

1. The solution as a YES or a NO. For example, "NO"

Present your final answer within $<$answer$>$ tags.

IMPORTANT: ONLY RETURN YES OR NO THAT IS THE ANSWER TO THE QUESTION. Do not include any explanations or additional information in your final answer.
\end{tcolorbox}
\newpage

\subsection{Example Prompt for Resolving Instability}

For the task of resolving instability in a given unstable solution, the prompt begins by providing models with the instance's preference profile (as with the prompts for the other tasks). In addition, the prompt includes an unstable matching, and asks the model to resolve the instability by outputting a stable solution (in an identical format to the prompt in \Cref{app:gen-stable-sol}).

\begin{tcolorbox}[fontupper=\sffamily]
    You are an intelligent assistant who is an expert in algorithms. Consider the following instance of the two-sided matching problem and respective unstable matching, where 5 men are to be matched with 5 women. 

Here are the preference lists for all individuals:

$<$preferences$>$

\{

M: \{

M1: [W5,W3,W4,W2,W1],

M2: [W2,W3,W5,W1,W4],

M3: [W5,W3,W1,W4,W2],

M4: [W1,W3,W2,W5,W4],

M5: [W2,W3,W4,W1,W5],

\},

W: \{

W1: [M3,M5,M4,M1,M2],

W2: [M1,M3,M4,M5,M2],

W3: [M3,M2,M4,M1,M5],

W4: [M4,M2,M3,M5,M1],

W5: [M2,M4,M5,M1,M3],

\}\}

$<$/preferences$>$

Here is an unstable matching.

$<$answer$>$

\{

        "M1": "W4",

        "M2": "W5",

        "M3": "W3",

        "M4": "W2",

        "M5": "W1"

\}

$<$/answer$>$

Your task is to modify the given unstable matching to make it equivalent to the proposer-optimal stable matching. You can use XML tags like $<$scratchpad$>$ to explain your thought process while computing the solution.

Once you have found a stable matching, please return your matching in the JSON format given below:

$<$answer$>$

\{

        "M1": "$<$woman matched with M1$>$",

        "M2": "$<$woman matched with M2$>$",

        "M3": "$<$woman matched with M3$>$",

        "M4": "$<$woman matched with M4$>$",

        "M5": "$<$woman matched with M5$>$"

\}

$<$/answer$>$

Make sure that each man/woman is matched with exactly ONE partner. It is mandatory that you provide a matching as a JSON object enclosed in $<$answer$>$$<$/answer$>$ tags as described above.
\end{tcolorbox}

\newpage
\subsection{Example Prompt for Repeated Queries Due to Missing JSON Object}\label{app:repeat-missing}

For tasks where the desired output is a JSON object (when outputting a stable solution), models are given an additional opportunity to rectify issues in their response if the original response is incorrectly formatted. The prompt below is passed to the model to help rectify issues related to missing JSON objects. Note that the $<$initially passed prompt$>$ and $<$last 3,000 characters of LLM's first response$>$ XML tags are replaced by the initial prompt and the tail of the models initial response, respectively.

\begin{tcolorbox}[fontupper=\sffamily]
    Previously, I gave you the following task:

---------------------------------------------------------

$<$initially passed prompt$>$

---------------------------------------------------------

In your response, you either failed to provide me with a matching or did not adhere to the JSON format I had asked for. Here are the last few lines of your response for reference:

---------------------------------------------------------

$<$last 3,000 characters of LLM's first response$>$

---------------------------------------------------------

Please correct your response and provide me with the matching in the following JSON format, enclosed in $<$answer$>$$<$/answer$>$ tags.$<$answer$>$

\{

        "M1": "$<$woman matched with M1$>$",

        "M2": "$<$woman matched with M2$>$",

        "M3": "$<$woman matched with M3$>$",

        "M4": "$<$woman matched with M4$>$",

        "M5": "$<$woman matched with M5$>$"

\}

$<$/answer$>$

Make sure that each man/woman is matched with exactly ONE partner.
\end{tcolorbox}

\subsection{Example Prompt for Repeated Queries Due to Incorrectly Formatted JSON Object}

Similar to the prompt in \Cref{app:repeat-missing}, the following prompt is passed to the model when the model's initial response contains an incorrectly formatted JSON object. Once again, the $<$initially passed prompt$>$ and $<$last 3,000 characters of LLM's first response$>$ XML tags are replaced by the initial prompt and the tail of the models initial response, respectively.

\begin{tcolorbox}[fontupper=\sffamily]
    Previously, I gave you the following task:

---------------------------------------------------------

$<$initially passed prompt$>$

---------------------------------------------------------

In your response, you failed adhere to the JSON format I had asked for. Here are the last few lines of your response for reference:

---------------------------------------------------------

$<$last 3,000 characters of LLM's first response$>$

---------------------------------------------------------

Please correct your response and provide me with the matching in the following JSON format, enclosed in $<$answer$>$$<$/answer$>$ tags.$<$answer$>$

\{

        "M1": "$<$woman matched with M1$>$",

        "M2": "$<$woman matched with M2$>$",

        "M3": "$<$woman matched with M3$>$",

        "M4": "$<$woman matched with M4$>$",

        "M5": "$<$woman matched with M5$>$"

\}

$<$/answer$>$

Make sure that each man/woman is matched with exactly ONE partner.
\end{tcolorbox}

\subsection{Example Prompt for Repeated Queries Due to Incomplete Matching}

Similar to the prompt in \Cref{app:repeat-missing}, the following prompt is passed to the model when the model's initial response contains a correctly formatted JSON object, but the matching itself is incomplete, or some agents have multiple partners. After the initially passed prompt, note that additional details are provided to assist the LLM in rectifying its response. Again, the $<$initially passed prompt$>$ and $<$last 3,000 characters of LLM's first response$>$ XML tags are replaced by the initial prompt and the tail of the models initial response, respectively.

\begin{tcolorbox}[fontupper=\sffamily]
    Previously, I gave you the following task:

---------------------------------------------------------

$<$initially passed prompt$>$

---------------------------------------------------------

In your response, the matching you selected involves some women being matched with multiple men, which is not allowed. For example, W2 is matched with M1, M2, and M5. Additionally, W3, and W4 are unmatched. Here are the last few lines of your response for reference:

---------------------------------------------------------

$<$last 3,000 characters of LLM's first response$>$

---------------------------------------------------------

Please correct your response and provide me with the matching in the following JSON format, enclosed in $<$answer$>$$<$/answer$>$ tags.$<$answer$>$

\{

        "M1": "$<$woman matched with M1$>$",

        "M2": "$<$woman matched with M2$>$",

        "M3": "$<$woman matched with M3$>$",

        "M4": "$<$woman matched with M4$>$",

        "M5": "$<$woman matched with M5$>$"

\}

$<$/answer$>$

Make sure that each man/woman is matched with exactly ONE partner.
\end{tcolorbox}

\section{Training Examples for Fine-tuning}\label{app:ft_seqs}

\subsection{System-prompt  (\texorpdfstring{$\boldsymbol{s}$}{s})}

This is the first part of the input, and is common across all tasks.

\begin{tcolorbox}[fontupper=\sffamily]
Below is an instruction that describes a task, paired with an input that provides further context. 

Write a response that appropriately completes the request. 

Before answering, think carefully about the question and create a step-by-step chain of thoughts to ensure a logical and accurate response.
\end{tcolorbox}

\subsection{High-level instruction (\texorpdfstring{$\boldsymbol{u}$}{u})}

\begin{itemize}
    \item \textbf{Generating:}

    \begin{tcolorbox}[fontupper=\sffamily]
    \#\#\# Instruction:
    
    You are an intelligent assistant who is an expert in algorithms. Your task is to find the proposer-optimal stable matching, for the two-sided matching problem.  

    \#\#\# Question:

    Consider the following instance of the two-sided matching problem, where 5 men are to be matched with 5 women.

    Here are the preference lists for all individuals:
    \end{tcolorbox}

    \item \textbf{Comprehension:}

    \begin{tcolorbox}[fontupper=\sffamily]
    \#\#\# Instruction:
    
    You are an intelligent assistant who is an expert in algorithms. You will be given an instance of the two-sided matching problem, and will be asked to answer a question about the preferences of the agents involved.  

    \#\#\# Question:

    First, here are the preference lists for all individuals:
    \end{tcolorbox}

\end{itemize}

\subsection{Preference Profile (\texorpdfstring{$\boldsymbol{p^{(i)}}$}{p(i)})}

\begin{tcolorbox}[fontupper=\sffamily]

    $<$preferences$>$

\{

M: \{

M1: [W5,W3,W4,W2,W1],

M2: [W2,W3,W5,W1,W4],

M3: [W5,W3,W1,W4,W2],

M4: [W1,W3,W2,W5,W4],

M5: [W2,W3,W4,W1,W5],

\},

W: \{

W1: [M3,M5,M4,M1,M2],

W2: [M1,M3,M4,M5,M2],

W3: [M3,M2,M4,M1,M5],

W4: [M4,M2,M3,M5,M1],

W5: [M2,M4,M5,M1,M3],

\}\}

$<$/preferences$>$
    
    $<$/preferences$>$
\end{tcolorbox}

\subsection{Task-prompt (\texorpdfstring{$\boldsymbol{t^{(i)}}$}{t})}

\begin{itemize}

    \item \textbf{Generating:}
    
    \begin{tcolorbox}[fontupper=\sffamily]
    Your task is to find the proposer-optimal stable matching.

    Once you have found a stable matching, please return your matching in the JSON format given below:

    $<$answer$>$
    
    \{
    
    	"M1": "$<$woman matched with M1$>$",

            "M2": "$<$woman matched with M2$>$",

        "M3": "$<$woman matched with M3$>$",

        "M4": "$<$woman matched with M4$>$",

        "M5": "$<$woman matched with M5$>$"
    
    \}
    
    $<$/answer$>$

    Make sure that each man/woman is matched with exactly ONE partner. It is important that you enclose your JSON object in $<$answer$>$$<$/answer$>$ tags.

    \end{tcolorbox}

    \item \textbf{Comprehension (Level-1):}

    \begin{tcolorbox}[fontupper=\sffamily]

    Now, you will be asked a specific question about agent preferences:

    $<$question$>$
    
    Who is agent W3's, 5-most preferred agent?
    
    $<$/question$>$

    Once you have determined the answer, provide your output in the following format:

    1. The solution as a single agent name. For example, "W1"

    Present your final answer within $<$answer$>$ tags.

    IMPORTANT: ONLY RETURN THE NAME OF THE SINGLE AGENT THAT IS THE ANSWER TO THE QUESTION. Do not include any explanations or additional information in your final answer.
        
    \end{tcolorbox}

    \item \textbf{Comprehension (Level-2):}

    \begin{tcolorbox}[fontupper=\sffamily]

    Now, you will be asked a specific question about agent preferences:

    $<$question$>$
    
    Would agent W3, prefer M4 and M3 over M1?
    
    $<$/question$>$

    Once you have determined the answer, provide your output in the following format:

    1. The solution as a YES or a NO. For example, "NO"

    Present your final answer within $<$answer$>$ tags.

    IMPORTANT: ONLY RETURN YES OR NO THAT IS THE ANSWER TO THE QUESTION. Do not include any explanations or additional information in your final answer.

    \end{tcolorbox}

    \item \textbf{Comprehension (Level-3):}

    \begin{tcolorbox}[fontupper=\sffamily]

    Now, you will be asked a specific question about agent preferences:

    $<$question$>$
    
    If agent W5 is currently engaged to M3, would she accept proposals from M1 and M4?
    
    $<$/question$>$

    Once you have determined the answer, provide your output in the following format:

    1. The solution as a YES or a NO. For example, "NO"

    Present your final answer within $<$answer$>$ tags.

    IMPORTANT: ONLY RETURN YES OR NO THAT IS THE ANSWER TO THE QUESTION. Do not include any explanations or additional information in your final answer.

    \end{tcolorbox}

\end{itemize}

The input $\mathbf x^{(i)}$ ends with the following text, to indicate the start of the chain-of-thought reasoning trace,

\begin{tcolorbox}[fontupper=\sffamily]
    \#\#\# Response:

    $<$think$>$
\end{tcolorbox}

\subsection{Reasoning-trace (\texorpdfstring{$\boldsymbol{r^{(i)}}$}{r}):}

\begin{itemize}
    \item \textbf{Generating:}

    \begin{tcolorbox}[fontupper=\sffamily]

    \#\#\# Response:

    $<$think$>$
    
    Okay, so I need to find a stable matching for 5 men and 5 women. For this, I can use the Gale-Shapley algorithm. Before I compute the solution for this instance, I will list the steps of the Gale-Shapley algorithm.

    1. Initialize all men and women as unmatched.
    
    2. Create a list to keep track of each man's next proposal (initially set to 0 for all men).
    
    3. While there are unmatched men:
    
       a. Select an unmatched man (M).
    
       b. Find the next woman (W) on M's preference list that he hasn't proposed to yet.
    
       c. If W is unmatched, match M and W.
    
       d. If W is matched but prefers M to her current partner:
    
          - Unmatch W from her current partner.
    
          - Match M and W.
    
          - Set the unmatched man as W's previous partner.
    
       e. If W rejects M, move to the next woman on M's preference list.
    
    4. Repeat step 3 until all men are matched.

    Next, I will exectute the above algorithm for the given instance. The steps are as follows:

    M1 is free. M1 proposes to W3
    
    Since W3 is free, W3 accepts the proposal. Now M1 and W3 are matched.
    
    M2 is free. M2 proposes to W5
    
    Since W5 is free, W5 accepts the proposal. Now M2 and W5 are matched.
    
    M3 is free. M3 proposes to W1
    
    Since W1 is free, W1 accepts the proposal. Now M3 and W1 are matched.
    
    M4 is free. M4 proposes to W4
    
    Since W4 is free, W4 accepts the proposal. Now M4 and W4 are matched.
    
    M5 is free. M5 proposes to W4
    
    Since W4 prefers their current partner M4 to M5, W4 rejects the proposal. M4 and W4 are still matched, and M5 is still free.
    
    M5 is free. M5 proposes to W1
    
    Since W1 prefers M5 to their current partner M3, W1 accepts the proposal. Now M5 and W1 are matched, and M3 is free.
    
    M3 is free. M3 proposes to W4
    
    Since W4 prefers their current partner M4 to M3, W4 rejects the proposal. M4 and W4 are still matched, and M3 is still free.
    
    M3 is free. M3 proposes to W3
    
    Since W3 prefers their current partner M1 to M3, W3 rejects the proposal. M1 and W3 are still matched, and M3 is still free.
    
    M3 is free. M3 proposes to W2
    
    Since W2 is free, W2 accepts the proposal. Now M3 and W2 are matched.
    
    $<$/think$>$
    \end{tcolorbox}

    \item \textbf{Comprehension (Level-1):}

    \begin{tcolorbox}[fontupper=\sffamily]

    Alright, I have been given the task of determining W3's 5-most preferred agent. For this, let me look at her preference list in order:

    position 1: M4
    
    position 2: M5
    
    position 3: M1
    
    position 4: M3
    
    position 5: M2

    Hence, W3's 5-most preferred agent is M2.
    
    $<$/think$>$
        
    \end{tcolorbox}

    \item \textbf{Comprehension (Level-2):}

    \begin{tcolorbox}[fontupper=\sffamily]
        Alright, I have been asked to determine whether agent W3 would prefer M4 and M3 over M1. For this, let me look at W3's preference list in order:
        
        position 1: M4
        
        position 2: M5
        
        position 3: M1
        
        position 4: M3
        
        position 5: M2

        So, M4 is W3's 1-most preferred partner, M3 is W3's 4-most preferred partner, and M1 is W3's 3-most preferred partner.

        Hence, agent W3 prefers M4 over M1, and agent W3 does not prefer M3 over M1. This means that the order of preference is M4 $>$ M1 $>$ M3.

        Therefore, since the question was whether agent W3 would prefer BOTH M4 AND M3, given that it DOES NOT prefer M3 over M1, the answer is 'NO'.
        
        $<$/think$>$

    \end{tcolorbox}

    \item \textbf{Comprehension (Level-3):}

    \begin{tcolorbox}[fontupper=\sffamily]

    Alright, I have been asked to determine whether agent W5 would accept a proposal from M1 or M4, if she is currently matched with M3. For this, let me look at W5's preference list in order:

    position 1: M5
    
    position 2: M4
    
    position 3: M2
    
    position 4: M3
    
    position 5: M1

    So, M1 is W5's 5-most preferred partner, M4 is W5's 2-most preferred partner, and M3 is W5's 4-most preferred partner.

    Hence, agent W5 does not prefer M1 over their current partner M3, and agent W5 prefers M4 over their current partner M3. This means that the order of preference is M4 $>$ M3 $>$ M1.

    Therefore, since the question was whether agent W5 would prefer EITHER M1 OR M4, given that it prefers M4 over M3, the answer is 'YES'.
    
    $<$/think$>$
            
    \end{tcolorbox}

\end{itemize}

\subsection{Answer (\texorpdfstring{$\boldsymbol{a^{(i)}}$}{a}):}

\begin{itemize}

    \item \textbf{Generating:}
    
    \begin{tcolorbox}
    [fontupper=\sffamily]
    
    $<$answer$>$
    
    \{
    
    	"M1": "W3",
    
            ...
    
    	"M5": "W1"
    
    \}
    
    $<$/answer$>$
    \end{tcolorbox}

    \item \textbf{Comprehension (Level-1):}

    \begin{tcolorbox}[fontupper=\sffamily]

    $<$answer$>$M2$<$/answer$>$
        
    \end{tcolorbox}

    \item \textbf{Comprehension (Level-2):}

    \begin{tcolorbox}[fontupper=\sffamily]

    $<$answer$>$NO$<$/answer$>$
        
    \end{tcolorbox}

    \item \textbf{Comprehension (Level-3):}

    \begin{tcolorbox}[fontupper=\sffamily]

    $<$answer$>$YES$<$/answer$>$
        
    \end{tcolorbox}

\end{itemize}

\end{document}